\documentclass[letterpaper, 10 pt, conference]{ieeeconf}

\IEEEoverridecommandlockouts                              % This command is only needed if 
\usepackage{times}
\usepackage{epsfig}
\usepackage{graphicx}
\usepackage{amsmath}
\usepackage{amssymb}
\usepackage{svg}
\usepackage{multirow}
\graphicspath{{figure/}}
\usepackage[font=small,skip=0pt]{caption}
\usepackage{algorithm}
\usepackage{algpseudocode}
\usepackage{amsmath}
\title{\LARGE \bf
Line-based Camera Pose Estimation in Point Cloud of Structured Environments
}

\author{Huai Yu$^{1,2}$, Weikun Zhen$^{2}$, Wen Yang$^{1}$ and Sebastian Scherer$^{2}$% <-this % stops a space
\thanks{$^{1}$Huai Yu and Wen Yang are with the Electronic Information School, Wuhan University,  Wuhan 430072, China {\tt\small \{yuhuai, yangwen\}@whu.edu.cn}}%
\thanks{$^{2}$ Weikun Zhen and Sebastian Scherer are with the Robotics Institute, Carnegie Mellon University,
        Pittsburgh, PA 15213, USA
        {\tt\small \{weikunz, basti\}@andrew.cmu.edu}}%
}

\begin{document}

\maketitle
\thispagestyle{empty}
\pagestyle{empty}

%%%%%%%%%%%%%%%%%%%%%%%%%%%%%%%%%%%%%%%%%%%%%%%%%%%%%%%%%%%%%%%%%%%%%%%%%%%%%%%%
\begin{abstract}
Accurate registration of 2D imagery with point clouds is a key technology for image-LiDAR point cloud fusion, camera to laser scanner calibration and camera localization. Despite continuous improvements, automatic registration of 2D and 3D data without using additional textured information still faces great challenges. In this paper, we propose a new 2D-3D registration method to estimate 2D-3D line feature correspondences and the camera pose in untextured point clouds of structured environments. Specifically, we first use geometric constraints between vanishing points and 3D parallel lines to compute all feasible camera rotations. Then, we utilize a hypothesis testing strategy to estimate the 2D-3D line correspondences and the translation vector. By checking the consistency with computed correspondences, the best rotation matrix can be found. Finally, the camera pose is further refined using non-linear optimization with all the 2D-3D line correspondences. The experimental results demonstrate the effectiveness of the proposed method on the synthetic and real dataset (outdoors and indoors) with repeated structures and rapid depth changes. %We will further release a 2D-3D registration dataset of structured environments.
\end{abstract}

%%%%%%%%%%%%%%%%%%%%%%%%%%%%%%%%%%%%%%%%%%%%%%%%%%%%%%%%%%%%%%%%%%%%%%%%%%%%%%%%
\section{Introduction}
The 2D to 3D registration problem is to match a query image with a 3D model to establish the geometric correspondence between the two modalities and estimate camera pose \cite{paudel2015local}. It is essential for many applications, e.g. point cloud colorization \cite{stamos2008integrating, dhall2017lidar} and camera localization \cite{campbell2018globally}. Based on the estimated transformation parameters, image textures can be used to colorize untextured point clouds, which is beneficial for further interpretation.   With the developments of low-cost LiDAR sensors and the advancements in LiDAR-based SLAM algorithms \cite{droeschel2018efficient, zhou2018automatic}, point cloud 3D models can be easily obtained. Thus 2D-3D registration can be used to localize a small lightweight camera inside pre-built 3D maps, which is attractive and complementary to existing visual SLAM technology.

%2D to 3D registration problem is to match a query image with a 3D model to establish the geometric correspondence between the two modalities and estimate camera pose \cite{paudel2015local}. It is important for information fusion \cite{stamos2008integrating}, camera to laser scanner calibration \cite{dhall2017lidar} and camera localization \cite{campbell2018globally}. With the developments of low-cost LiDAR sensors and the advancements in LiDAR-based SLAM algorithms \cite{droeschel2018efficient}, the untextured and unorganized point cloud 3D model can be easily obtained. However, without textured information, it is difficult to recognize semantic information even for artificial visual interpretation. A common practice is to rigidly couple a calibrated camera \cite{dhall2017lidar,zhou2018automatic} to colorize the point clouds. Nevertheless, inaccurate extrinsic calibration of the two sensors may produce unexpected effects on the colorization. Additionally, localizing a small lightweight camera inside pre-built 3D maps is attractive and complementary to existing visual SLAM technology.  %Moreover, the rigid bundling sometimes restricts the freedom of the two sensors.  Thus efficient registration of imagery with point clouds is essential and useful to the community.

However, the 2D to 3D registration is very challenging because of the appearance differences and modality gaps. Generally, the current 2D to 3D matching methods are often established on the same kind of descriptors (e.g. SIFT) across different modalities \cite{ sattler2017efficient}. Nevertheless, appearance and visual feature changes may occur between viewpoints, light conditions, weather and seasons, which make visual features not suitable for the registration of optical imagery with point clouds. Additionally, the same kinds of appearances and visual features are not always available for point cloud data.  %Although there are many works for camera to point cloud registration, they are either based on additive information like textures \cite{sattler2011fast, sattler2017efficient}, or using an interactive method \cite{sun2002interactive, liu2012systematic}. 
On the other hand, LiDAR point clouds are 3D data with geometric position, while images are projected 2D data with textured information. The modality differences and description gaps make the common image registration methods fail to get correspondences. Fortunately, 2D images share some geometric consistent features with point clouds, such as line segments and planes \cite{goto2018line}. Thus we can use geometric co-occurrence to find these feature correspondences. We focus on the scenarios with rich geometric information, such as urban scene with buildings. This kind of Manhattan world commonly consists of a triplet of 3D parallel lines, which are shown as 2D lines intersecting at vanishing points in image planes \cite{lezama2014finding}. These geometric constraints are beneficial for establishing the correspondences between the two modalities.

 {\color{black} The global 2D-3D registration is known as a difficult problem for the non-convexity and feature description gap, it is very vital to robotics because of drift-free localization. However, there is a lack of research on globally registering a single image to a point cloud map. In this paper, we propose a global 2D-3D registration pipeline to estimate line correspondences and the camera pose for structured environments. By discovering the geometric relationship of 2D-3D lines, we propose to match primary vanishing directions with 3D parallel orientations to decouple the camera rotation and position estimations. Then we further use a simple hypothesis testing strategy to eliminate the rotation ambiguities and simultaneously estimate the translation vector and the 2D-3D line correspondences. Compared with existing methods, our proposed method globally estimates the camera pose without any pose initialization or texture information. The pipeline of using vanishing direction matching and hypothesis testing gives more reliable 2D-3D correspondences, additionally can well avoid the camera pose estimation being stuck into a local optimum. It is also robust to feature outliers and can deal with challenging scenes with repeated structures and rapid depth changes.}

The structure of this paper is organized as follows: Section \ref{sec2} reviews previous work related to 2D-3D registration. Section \ref{sec3} details the methodology. Experimental results and discussions are presented in Section \ref{sec4}. Finally, Section \ref{sec5} gives the conclusion of this paper. 
\begin{figure*}[t]
	\begin{center}
		\includegraphics[width=0.99\linewidth]{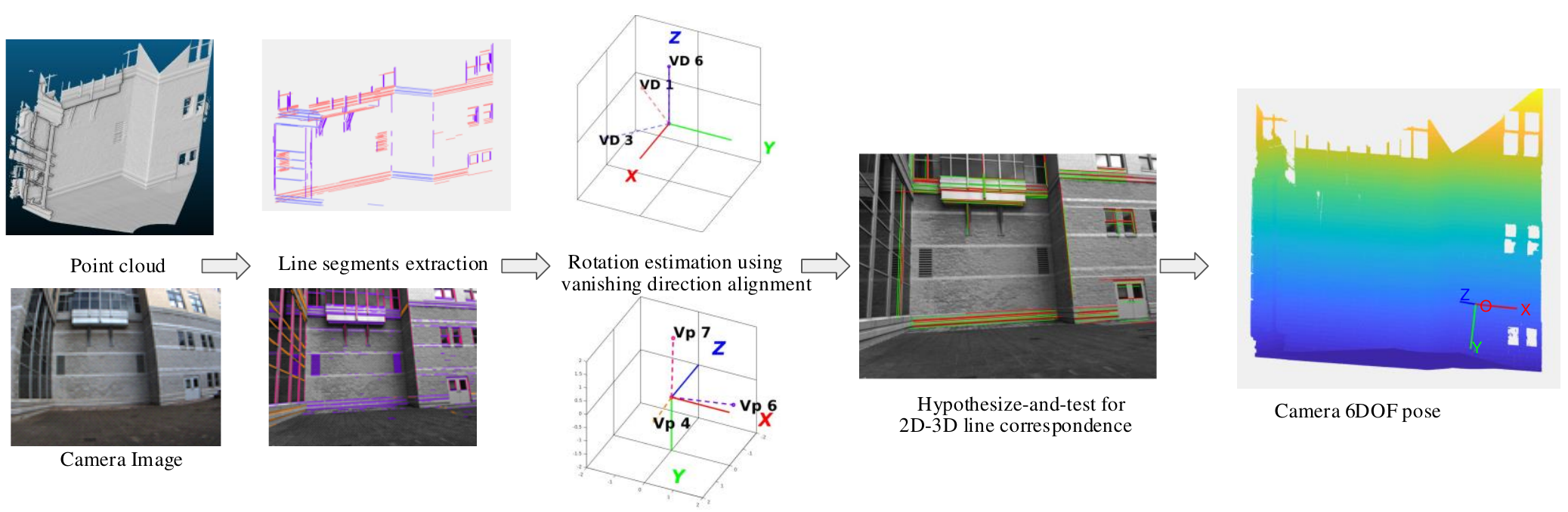}
	\end{center}
	\caption{Pipeline of 2D-3D registration for line correspondence and pose estimation.}
	\label{fig:overview}
\end{figure*}
%or only get a local optima \cite{david2004softposit, moreno2008pose}. 
%------------------------------------------------------------------------
\section{Related Work}
\label{sec2}

For image to point cloud registration, the general approach is transferring one modality to the other, i.e. project the point cloud to image space \cite{feng2016fast,sattler2017efficient} and reconstruct the point cloud using multi-view geometry \cite{stamos2008integrating}, and then registering at the same dimension. However, the reconstruction approaches do not work well for the registration of a single image to point clouds. For the projection approaches, because there is no texture information in the point cloud, geometric features are often used, which are more robust than appearance features. Therefore, the main issue is how to estimate 2D-3D feature correspondences and the camera pose.

With 2D-3D feature correspondences, the pose estimation problem can be well solved by PnP or PnL algorithms \cite{lepetit2009epnp,xu2017pose}. However, in our cases, it is very challenging because both the pose and correspondence are unknown,  which is considered a chicken-and-egg conundrum. A conventional approach is using the RANSAC-based strategy \cite{fischler1981random}, which can simultaneously conider feature correspondence and geometric information. However, without any constraint, RANSAC suffers from a large search space and high time complexity to avoid the local optimum. To guarantee global optimum and improve the efficiency, the {\color{black}existing} approaches mainly rely on point and line feature correspondence  \cite{campbell2018globally, david2004softposit,kamgar2011matching, liu2012systematic,pumarola2017pl}. Several approaches simultaneously determine the correspondence and pose between 2D and 3D points, such as SoftPosit \cite{david2004softposit} and BlindPnP \cite{moreno2008pose}. {\color{black}However, they are local optimization methods requiring good pose initializations. And camera to LiDAR sensor calibration methods \cite{levinson2013automatic,zhou2018automatic} also optimize the extrinsic transformation based on a very reliable pose initialization. Besides, recent 2D-3D tracking methods \cite{marchand1999robust,zuo2019visual} use the pose of previous frame to predict the current frame pose to implement 2D-3D matching, which also belong to the local optimization methods.}  More recent works use branch-and-bound strategies to guarantee the global optimum without a pose prior \cite{brown2019family,campbell2018globally}. However, these approaches start from the existing point features and are conducted on synthetic data which guarantees the proportion of inlier correspondences, which are very difficult to follow for real data. 

For real 2D-3D data, it is troublesome to extract the highly repeatable features across modalities. Point features are often used (e.g. Harris, SIFT, junctions \cite{xia2014accurate}), but they need careful design to encode geometric information and maintain the co-occurrence ability across modalities. {\color{black}Compared with these point features}, line features are more suitable for 2D-3D registration, which share the characteristics of stability and representative. However, it is difficult to describe the local information of line feature for both 2D and 3D data. An early approach modifies SoftPosit algorithm for line feature \cite{david2003simultaneous}. However, it needs a good initialization and may get stuck in a local optimum. Recently, Brown et al. \cite{brown2019family} utilize both point features and line features to minimize the projection error, then use branch-and-bound formulation to guarantee the global optimization. 

Structured environments like buildings are different to natural scenes because of the repeated structure and weak texture. The aforementioned methods may fail to find a good pose. Fortunately, many 3D parallel lines in structured environments generate vanishing points in 2D images. In previous work \cite{lee2015real, antone2000automatic}, vanishing points have been used for rotation estimation between images, but rarely used in registration between 2D and 3D data. A systematic 2D-3D registration method is proposed for 2D-image and 3D-range in an interactive manner \cite{liu2012systematic}. The camera orientation is recovered by matching vanishing points with 3D directions and translation is estimated by RANSAC among all the linear features. Because it is hard to determine the correspondence between vanishing points with 3D directions, the authors interactively rotate the 3D model to align 3D directions with 2D vanishing directions. Therefore, there is a great demand for exploring the automatic 2D-3D registration problem of data in structured environments.
%In this paper, we rebuild the approach based on primary vanishing directions (extracted from 2D image) and 3D parallel line orientations. A new way is designed to extract 2D and 3D line segments, calculate rotation candidates, estimate line correspondence and translation vector. Most importantly, it is unnecessary to match 3D orientations with 2D vanishing directions in an interactive way. The rotation ambiguity is well solved by searching the largest number of inliers among different rotation candidates. It should be noticed that the proposed method works with the precondition of a large overlap between point clouds and images, which is consistent with the condition of previous methods.

\section{Overview of the proposed method}
\label{sec3}

To address the automatic registration of a single image with point clouds in structured environments, we propose to separately estimate the rotation matrix and translation vector by using geometric correspondences. It starts from a single image and untextured point clouds, and outputs the 2D-3D line correspondences and camera poses related to the point cloud frame. We first extract line segments from both 2D and 3D data and cluster them into two sets (vanishing point lines for 2D and parallel lines for 3D). Based on the relationship of vanishing points and the parallel directions of 3D lines \cite{lee2015real}, we then coarsely match primary 2D vanishing directions and 3D parallel orientations to compute feasible rotation matrices. With each rotation matrix candidate, we use a hypothesis testing strategy to estimate the translation vector. The estimation with the most line correspondence inliers is {\color{black}a reliable initialization of camera position}. Additionally, by comparing the number of inliers for different rotation candidates, we can estimate the final line correspondences and translation vector, simultaneously removing the wrong rotation estimations. After obtaining the 2D-3D line correspondences we further optimize the camera pose by minimizing the line projection error.  The main steps include line segment extraction, rotation matrix candidate estimation, line correspondence estimation, and final pose refinement using line correspondence (Fig.\ref{fig:overview}).

\subsection{Line segment extraction for image and point cloud}
\begin{figure}[htp]
	\centering
	\begin{minipage}[b]{0.49\linewidth}
		\centering
		\includegraphics[width = .9\columnwidth]{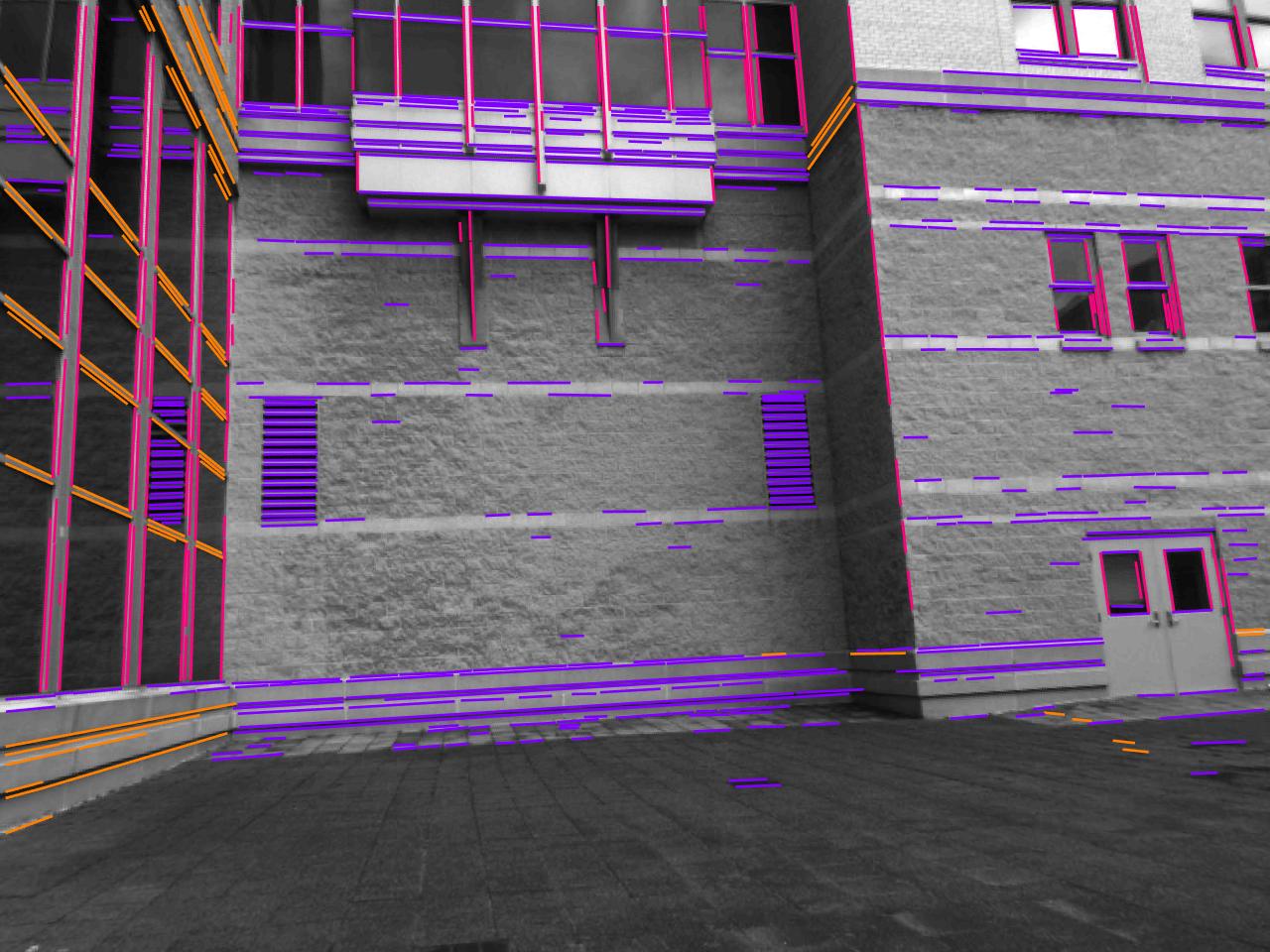} 
		\centerline{\scriptsize(\textbf{a}) 2D line detection } 
		\label{fig:2dline}
	\end{minipage}
	\begin{minipage}[b]{0.49\linewidth}
		\centering
		\includegraphics[width = .8\columnwidth]{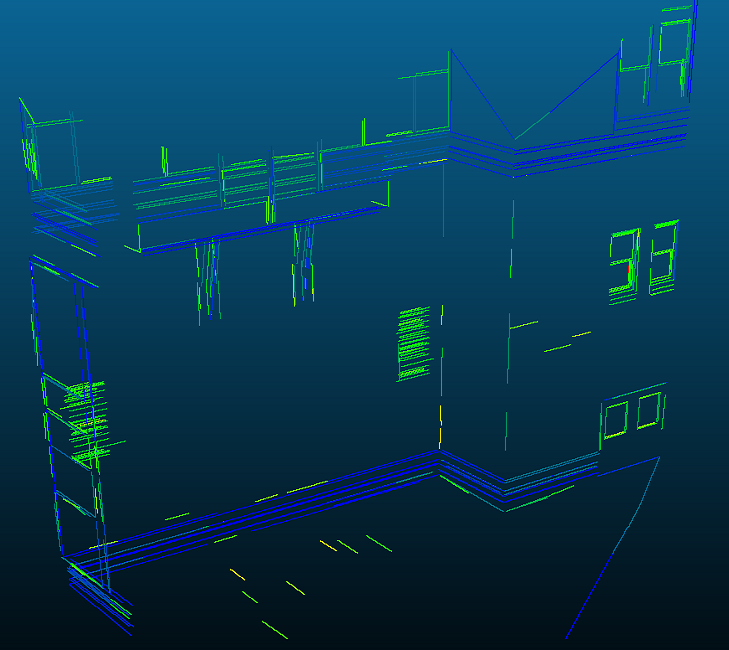}
		\centerline{\scriptsize(\textbf{b}) 3D line detection} 
		\label{fig:3dline}
	\end{minipage}
	\caption{A demonstration of the 2D and 3D line segment extraction method.}
	\label{fig:lineseg}
\end{figure}

For line segment extraction of images, many state-of-the-art methods exist in literature \cite{von2008lsd,xue2018learning}. Considering the computation efficiency and quality of extracted lines, we choose the LSD method \cite{von2008lsd}. However, the line segment extraction methods for unorganized point clouds are specified by the geometric structures. Here, we utilize a simple and efficient 3D line detection algorithm \cite{lu2019fast}, which is very suitable for the structured environments with the existence of many plane features. Fig.\ref{fig:lineseg} shows an example of the line segment extraction results for both 2D and 3D data. Their structures are very similar and share many co-occurring line segments, which is very important for later correspondence estimation.
%Finally, the adjacent 3D line segments are merged to obtain stable and significant ones.  

%We first segment the point clouds into 3D planes via normal clustering and region growing. By building the K nearest neighbor (KNN) tree structure, the normal of each point is estimated based on Principal Component Analysis (PCA) of the neighboring surface. Then we cluster the points with parallel normal directions in KNN tree structure. The adjacent points are co-planar if a) their normal directions are parallel; and b) the connected line of two points is orthogonal to the normal. After that, many small co-planar regions are obtained. We repeat the plane merging procedure one more time on the small regions to make it more stable. For the neighboring regions that cannot be merged, the adjacent points belonging to different regions are regarded as boundary points. Then each 3D plane region is projected to a 2D plane using orthographic projection with a standard perspective camera model. We can also use LSD to get the line segments for the 2D planes. However, because the projected 2D planes are with determined boundaries, we use contour detection and square line fitting to extract line segments, which are efficient and stable. Finally, the 2D line segments are re-projected to the point clouds to obtain the 3D lines.

\subsection{Rotation matrix candidate estimation}
\begin{figure}[htbp]
	\begin{center}
		\includegraphics[width=0.99\linewidth]{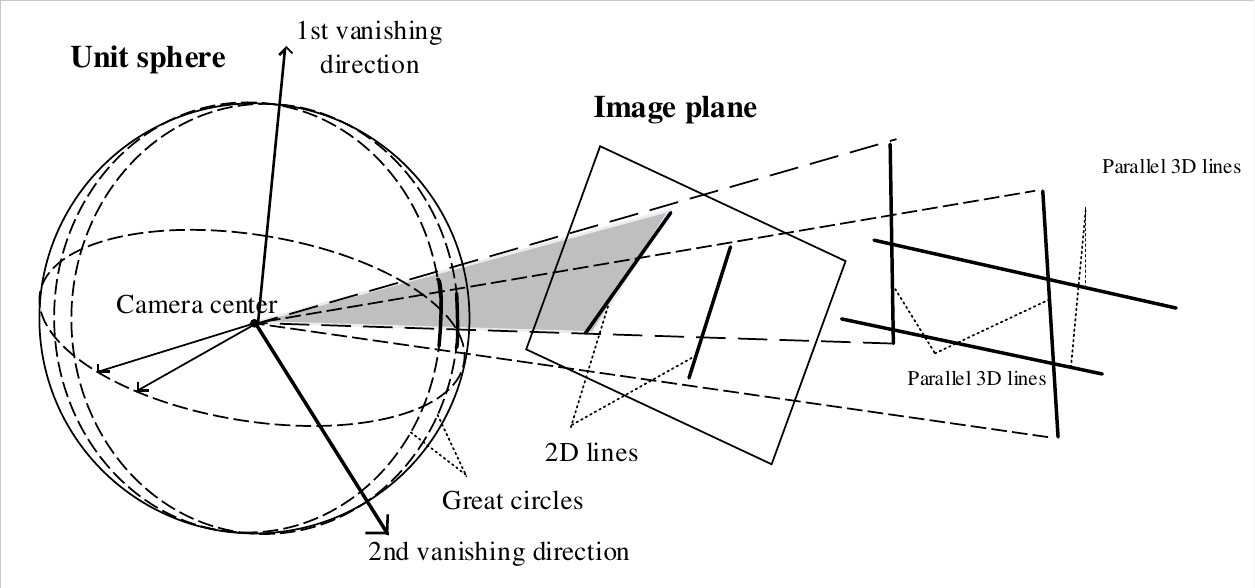}
	\end{center}
	\caption{Projection geometry of 3D parallel line segments}
	\label{fig:vanishing}
\end{figure}
The projections of 3D parallel line segments are 2D line segments sharing with the same vanishing point. As shown in Fig.\ref{fig:vanishing}, there is a set of 3D parallel lines with normalized homogeneous orientation $\widetilde{v}_{3d}=\left[\begin{matrix}V_x&V_y&V_z &0\end{matrix} \right]^T$ in the point cloud frame. Their projections on the unit sphere in camera frame are great circles intersecting at one point on the sphere. The direction from camera center to the point $\widetilde{v}_{2d}=\left[\begin{matrix}v_x&v_y&v_z& 0\end{matrix} \right]^T$ forms a vanishing direction which is corresponding to the parallel 3D lines in camera frame. Thus, the transformation from $\widetilde{v}_{3d}$ to $\widetilde{v}_{2d}$ is a typical Euclidean 3D transform
\begin{equation}
\begin{aligned}
\widetilde{v}_{2d}&=\left[\begin{matrix}R & t\end{matrix}\right]\cdot \widetilde{v}_{3d}  \\ 
v_{2d}&=R\cdot v_{3d}
\end{aligned}
\end{equation}
where $R$ and $t$ are the rotation and translation parameters from point cloud coordinate system to camera coordinate system, $v_{2d}$ and $v_{3d}$ denote the inhomogeneous unit vanishing directions and 3D orientations. We can observe that the rotation is totally determined by the direction correspondences. The rotation matrix can be estimated with at least 2 correspondences. 
\begin{equation} \label{eq2}
\left[\begin{matrix}v_{2d}^1& v_{2d}^2 &v_{2d}^3 \end{matrix}\right]=R \cdot \left[\begin{matrix}
v_{3d}^1 & v_{3d}^2 & v_{3d}^3
\end{matrix}\right]
\end{equation}
where $v_{2d}^1$ is corresponding to $v_{3d}^1$, $v_{2d}^2$ is corresponding to $v_{3d}^2$, respectively. $v_{2d}^3$ and $v_{3d}^3$ can be the cross products of the former two orientations or another correspondence. {\color{black} It should be noticed that the vanishing directions are rays while 3D line orientations are direction ambiguous. Thus with two correspondences we can get four rotation matrices. We first obtain several matrices using $M=v_{2d}*v_{3d}^{-1}$. For each estimation $M$, we need further find the orthonormal matrix $R$ to represent the rotation matrix.}

%For a single line segment, the cross product of the start point and the end point in camera frame forms the normal of the plane (crossing the line and the projection center) with the precomputed camera matrix $K$.
From the extracted 3D and 2D line segments, we need to cluster at least two primary 2D vanishing directions and two corresponding 3D directions. For 2D lines, we use a multiple RANSAC-based vanishing direction detector \cite{rother2002new} to cluster the lines into several sets. {\color{black}After clustering, vanishing direction is computed using PCA. SVD decomposition is utilized on the normals of the great circles of these lines (Fig.\ref{fig:vanishing}), the eigenvector with the smallest eigenvalue is the vanishing direction.} Likewise, we cluster 3D parallel lines in the point cloud frame using RANSAC-based detector. We randomly select one 3D line segment, clustering the other 3D lines with similar normalized 3D directions ($ < 1^\circ$) to get the largest number of inliers. 
%We randomly select two line segments, the cross product of the two normals forms a vanishing direction candidate. Then this candidate is tested on the rest line segments. The vanishing directions with the largest vote cluster several 2D line sets.
To maintain the robustness of the vanishing direction and 3D orientation estimation, the number of 2D and 3D clusters (denoted as $c$) needs to be set more than 3, e.g. c=5. After getting $c$ directions, we further merge the line sets with collinear and adjacent directions {\color{black}as \cite{liu2012systematic}}  and select the former two vectors with largest number of lines as the final 2D and 3D directions. Although there may exist more than two primary sets of line segments in 2D and 3D data, we only use the former two or three ones because it is already very stable for structured data.

  However, the geometric distribution of line segment sets is not robust enough to distinguish the orientation correspondences between 2D-3D line segment sets. {\color{black}Giving two pairs of vanishing directions and 3D orientations, 2 possible correspondences can be obtained. Additionally, the ambiguity of 3D orientations (two opposite directions) gives 4 possibilities of rotation estimation. Thus there will be 8 rotation candidates.} For three sets of 2D-3D line correspondences, the additional correspondence can be used for validation of the estimation. Thus there will be less than eight candidates. The rotation candidate estimation algorithm is outlined in Algorithm \ref{alg:Framwork}. %For two vanishing directions with $v_{2d}=(v_{2d}^1,v_{2d}^2)$ and two 3D orientations with $v_{3d}=((-1)^a*v_{3d}^1,(-1)^b*v_{3d}^2)$, $a,b\in{1,2}$, the rotation estimation can be $\mathcal{R}=\{R_1, R_2...,R_8\}$.
\begin{algorithm}[htb]
	\caption{ Rotation matrix candidate estimation}
	\label{alg:Framwork}
	\begin{algorithmic}[1]
		\Require
		2D line segments $\{l_{2d}\}$, 3D line segments $\{l_{3d}\}$;
		\Ensure
		8 rotation candidates $R=\{R_1, R_2...,R_8\}$;
		\State Clustering 2D lines into $M_{2d} (M_{2d}>3)$ sets and calculating the $M_{2d}$ vanishing directions;
		\label{code:fram:extract}
		\State Merging the $M_{2d}$ vanishing directions and picking the former 2 with most number of line segments $v_{2d}=\left( v_{2d}^1, v_{2d}^2 \right)$;
		\label{code:fram:trainbase}
		\State Clustering 3D lines into $M_{3d} (M_{3d}>3)$ parallel line sets and calculating the $M_{3d}$ 3D orientations;
		\label{code:fram:add}
		\State Merging the $M_{3d}$ 3D directions and picking the former 2 with most number of 3D line segments {\color{black}$v_{3d}=\left((\pm 1)*v_{3d}^a,(\pm 1)*v_{3d}^b\right), a,b \in\{1,2\}, ab=2$};
		\label{code:fram:classify}
		\State Calculating the rotation matrices using Eq.\ref{eq2}, $\mathcal{R}=\{R_1, R_2...,R_8\}$;
		\label{code:fram:select} \\
		\Return $\mathcal{R}$;
	\end{algorithmic}
\end{algorithm}
% merging part is not here
%After getting the original 2D and 3D line segments,  %For each segment, we first compute the distance between its center point to all other segments and discard the far ones (0.6 pixels for 2D lines, 0.06m for 3D lines). Then the clustered ones are further merged to a new line segment when their closest end point distance is small, i.e., extension or overlap (60 pixels for 2D lines, 0.6m for 3D lines). Finally, the isolated short line segments (80 pixels for 2D lines, 1.2m for 3D lines) are removed to get the final 2D and 3D segments.
\subsection{Translation vector and line correspondence estimation}
\begin{figure}[tbp]
	\begin{center}
		\includegraphics[width=0.6\linewidth]{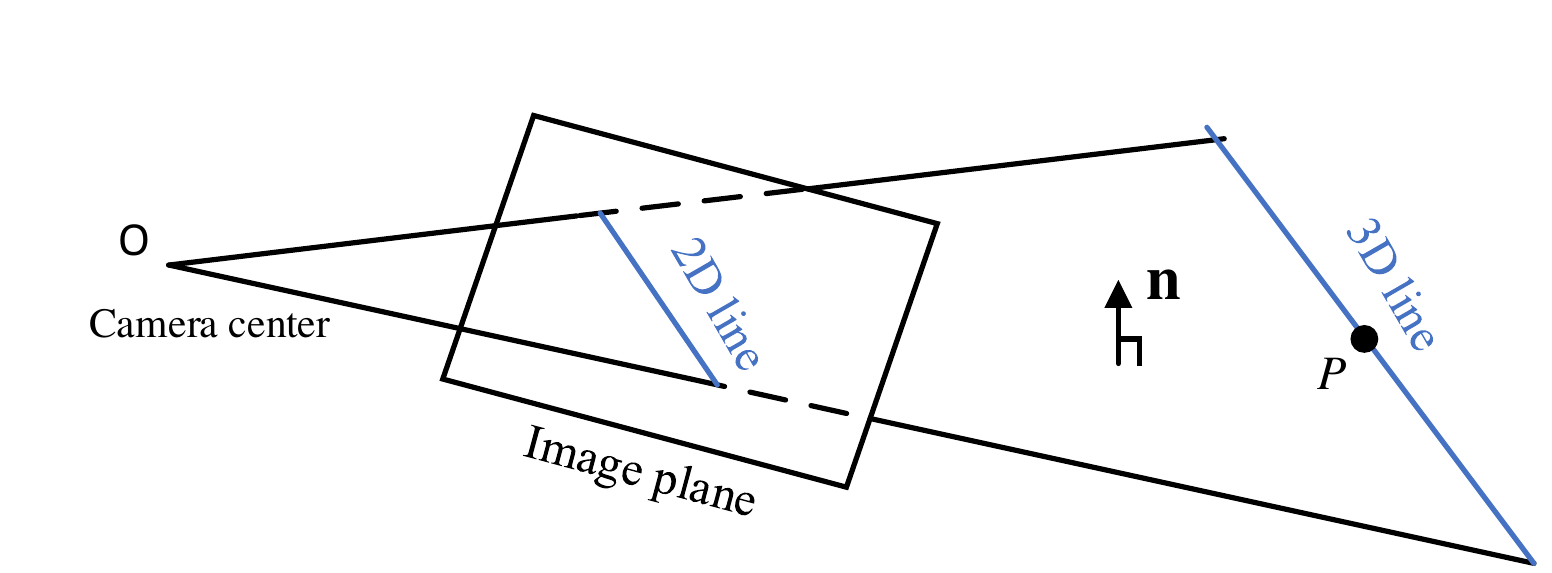}
	\end{center}
	\caption{Geometry between a 2D line and the corresponding 3D line.}
	\label{fig:coplanar}
\end{figure}

Although there exists ambiguities of the rotation estimation, the searching space is greatly decreased. For each rotation estimation candidate, the correspondence of 2D line sets with 3D line sets is determined. By analyzing the projection model from point cloud frame to camera frame, it is still a non-linear and non-convex problem. Bad initialization of the position will result in local optimum. Thus a hypothesis testing method on individual line correspondence can be used to decrease the possibility of getting stuck in a local optimum since there are only three translation parameters left. Before picking 2D-3D line correspondences, some isolated short segments are removed for better efficiency. For each 2D-3D line segment correspondence, the transformed 3D line segment in camera frame is co-planar with the 2D line segments (Fig.\ref{fig:coplanar}), thus the transformation of the 3D line center point $P$ is perpendicular to the normal of the corresponding 2D line segment $n$,
\begin{equation} \label{coplanar}
(RP+t)\cdot n=0.
\end{equation}

By randomly selecting 3 pairs of 2D-3D line correspondences, a translation vector can be calculated. Then the estimated $R$ and $t$ are used to estimate the total inlier correspondences. Each 3D line is first transformed to the camera coordinate system. For each 2D segment, several co-planar 3D segments are selected using Eq.\ref{coplanar} and are projected to the image plane as finite 2D lines. We calculate the overlap length and select the inliers with more than $50\%$ of 2D segment length. This constraint using overlap avoids the degenerate case where all 3D line segments become inliers when the camera is sufficiently distant. For each RANSAC iteration, we can get a translation estimation and a certain number of 2D-3D line correspondences. When the number of inliers exceeds $70\%$ of the total number of 2D (3D) line segments, or the iteration time reaches the setting maximum, it returns the estimation with the largest number of inliers as the estimated translation.
Thus, we can get 8 translation estimations for 8 rotation candidates, respectively. Then the one with the most inliers is selected as the final translation vector. At the same time, we obtain pairwise 2D-3D line correspondences and eliminate the rotation ambiguities. The hypothesis testing strategy takes the geometric distribution and individual line segment correspondence into consideration, which greatly decreases the possibility of being stuck into local optimum. {\color{black}The pseudocode of translation estimation part} is shown in Algorithm \ref{alg:trans}.

\begin{algorithm}[htbp] \color{black}{
	\caption{ Translation and line correspondence estimation}
	\label{alg:trans}
	\begin{algorithmic}[1]
		\Require
		2D line segments $\{l_{2d}\}$, 3D line segments $\{l_{3d}\}$ and 8 rotation candidates $\mathcal{R}$;
		\Ensure
		Optimal rotation $R^*$, translation $t^*$ and 2D-3D line correspondences $V_{2d-3d}$;
		
		\State Merging 2D and 3D line segments and removing short isolated segments;
		
		\ForAll {$R_i$ $\in$ $\mathcal{R}$}
		\State Initializing number of correspondences $N_i=0$;
		\Loop 
		\If {$N_i>0.7\times N(l_{2d})$ or max iterations} \indent ~~ terminate;
		\EndIf
		\State Randomly matching 3 pairs of 2D-3D segments, \indent ~~ using Eq.\ref{coplanar} to compute $t$;
		\State Calculating line overlap length and counting \indent ~~ the number $N$ of matching line sets $V$ with line \indent ~~ overlap length ($>0.5\times length(l_{2d})$);
		\If {$N>N_i$} $N_i=N, V_{2d-3d}^i=V, t_i=t$; 
		\EndIf
		\EndLoop
		\EndFor
		\State Picking $N^*=\max\{N_1, N_2,..., N_8\}$ and the corresponding $R^*, t^*, V_{2d-3d}$;
		 \\
		\Return $R^*$, $t^*$ and $V_{2d-3d}$;
	\end{algorithmic}}
\end{algorithm}

\subsection{Pose refinement using line correspondence}
\label{localrefine}
At the former stage, we obtain pairwise line correspondences and 6DOF pose parameters. Similar to point-based registration methods \cite{brown2007automatic},  we further optimize the camera pose by minimizing the projection error of all correspondences. However, it is not easy to use Euclidean distance to measure the projection error for line correspondences. For a pair of matching 2D-3D line segments, the registration error contains overlap distance and angle difference. Because the overlap length has been constrained at the inlier estimation step, we further optimize it with the collinear constraint. If the projections of two 3D-line end points are collinear to the corresponding 2D lines, there will be no angle difference between the correspondence. For two 3D end points $P_i=(P_i^p, P_i^q) $, the variables are camera pose $R,t$, {\color{black}its Lie algebra is $\xi$}. The projected end points are $u_i=(u_i^p, u_i^q)$, 
\begin{equation}
u_i \sim Kexp(\xi) P_i.
\end{equation}
We want to minimize the distance of both end points to the corresponding infinite 2D line $Ax+By+C=0$, whose coefficient vector is denoted as {\color{black}$H=\left[\begin{matrix}A&B&C\end{matrix} \right]$. Considering all the matching 2D-3D line segments, the minimization function can be formulated as
\begin{equation} \label{costs}
\begin{aligned}
\xi^{\star}&=\arg \min_{\xi}\sum_{i=1}^{N}e_i \\
&=\arg \min_{\xi} \frac{1}{2} \sum_{i=1}^{N}\frac{||H\cdot K \exp(\xi) P_i||^2_2}{A^2+B^2},
\end{aligned}
\end{equation}}
where $P_i$ contains two end points, the $L_2$ distance is the sum of the end points to the corresponding infinite line. It is finally formulated as a non-linear least squares problem. With Lie algebra, we can transform it to the unconstrained optimization problem and use the L-M algorithm to solve it. For a 3D end point $P$, the 3D transformed point $P'=\left[\begin{matrix}X'&Y'&Z'\end{matrix} \right]^T=RP+t$, the projected 2D point is $u=[u_x, u_y]^T$. The Jacobian matrix of the cost function is
\begin{equation} 
\frac{\partial e}{\partial \delta\xi}=\frac{\partial e}{\partial u}\frac{\partial u}{\partial P'}\frac{\partial P'}{\partial \delta\xi} ,
\end{equation}
where $\frac{\partial e}{\partial u}$ is the partial derivative of a 2D point to a 2D line,  
\begin{equation}
\frac{\partial e}{\partial u}=\left[\begin{matrix}\frac{\partial e}{\partial u_x} & \frac{\partial e}{\partial u_y} \end{matrix} \right]= \left[\begin{matrix}\frac{A}{\sqrt{A^2+B^2}} &
		\frac{B}{\sqrt{A^2+B^2}} \end{matrix} \right],
\end{equation}
and $\frac{\partial u}{\partial P'}\frac{\partial P'}{\partial \delta\xi}$ is the standard 3D to 2D projection model \cite{hartley2003multiple}, 
\begin{equation}
\frac{\partial u}{\partial P'}\frac{\partial P'}{\partial \delta\xi}=\left[\begin{smallmatrix}
\frac{f_x}{Z'} & 0& -\frac{f_xX'}{Z'^2} & -\frac{f_xX'Y'}{Z'^2} & f_x+\frac{f_xX'^2}{Z'^2} & -\frac{f_xY'}{Z'} \\
0 & \frac{f_y}{Z'}& -\frac{f_yY'}{Z'^2} & -f_y-\frac{f_yY'^2}{Z'^2} & \frac{f_yX'Y'}{Z'^2} & \frac{f_yX'}{Z'}
\end{smallmatrix}  
\right].
\end{equation}
We can use g2o library \cite{kummerle2011g} to implement the optimization (Eq.\ref{costs}). To remove outliers, we iteratively optimize the cost function and reject the outliers using the refined pose. The iteration terminates when there is no outliers or maximum iterations reached. 

\section{Experiments} 
\label{sec4}
To demonstrate the effectiveness of the proposed method, we test it on both synthetic (\ref{synthetic}) and real data (\ref{real}). {\color{black}Both data are structured environments including 3D parallel lines, which is the prerequisite for the vanishing direction matching.} It should be noticed that the proposed method works with the assumption of a large overlap between point clouds and images, which is consistent with the condition of previous methods \cite{campbell2018globally,goto2018line,liu2012systematic}.

\subsection{Synthetic data experiments} \label{synthetic}
\begin{figure}[htp]
	\centering
	\begin{minipage}[b]{0.49\linewidth}
		\centering
		\includegraphics[width = .9\columnwidth]{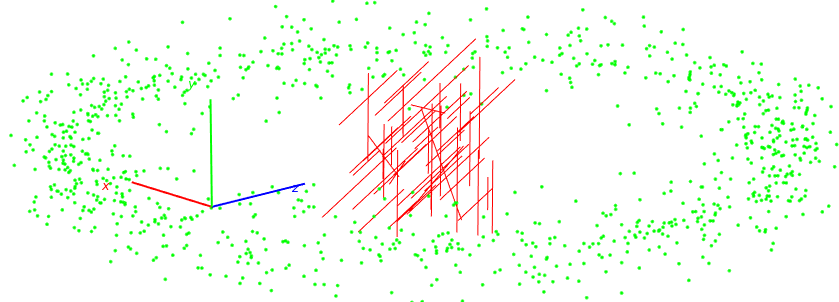} 
		\centerline{\scriptsize(\textbf{a}) 3D Result } 
		\label{fig:3dr}
	\end{minipage}
	\begin{minipage}[b]{0.49\linewidth}
		\centering
		\includegraphics[width = .9\columnwidth]{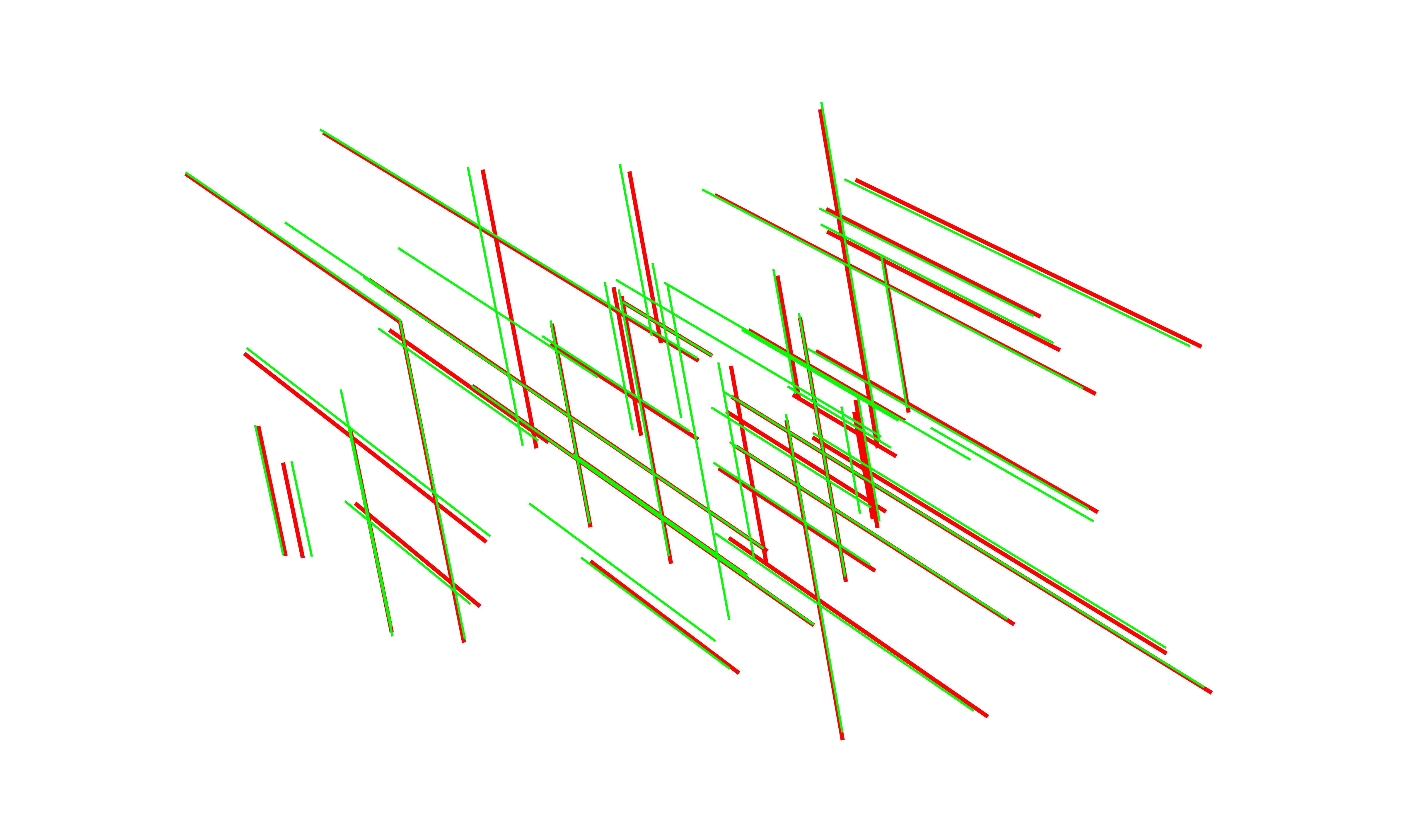}
		\centerline{\scriptsize(\textbf{b}) 2D Result} 
		\label{fig:2dr}
	\end{minipage}
	\caption{Synthetic 3D and 2D experimental results using random 3D lines. (a) 3D models(red lines), generated pose priors(green points) and estimated camera pose (o-xyz). (b)2D projection alignment results. 3D line projections shown as green lines, red as 2D image lines. }
	\label{fig::simulations}
\end{figure}

To evaluate the proposed method with a setting where the true camera pose was known, 50 independent Monte Carlo simulations are conducted. Two sets of random 3D parallel lines are generated {\color{black}from $[-1,+1]^3\in \mathbf{R}^3$}, a fraction of 3D lines with random orientations are added to form the original 3D line segments; a fraction of the 3D lines are randomly selected as outliers to model occlusion; the inliers are projected to a $640\times 480$ pixels virtual image {\color{black}with a synthetic camera $f_x=f_y=800$}; Gaussian noise is added to the 2D line endpoints with a standard deviation $\sigma$ of 2 pixels; and some 2D lines with random orientation are added to the image as the 2D line outliers. Based on these setups, we do not need to conduct line segment extraction for images and point clouds. Visualization of synthetic setups and registration results are shown in Fig.\ref{fig::simulations}. 

\begin{figure}[hbtp]
	\centering
	\begin{minipage}[b]{0.32\columnwidth}
		\centering
		\includegraphics[width = .99\columnwidth]{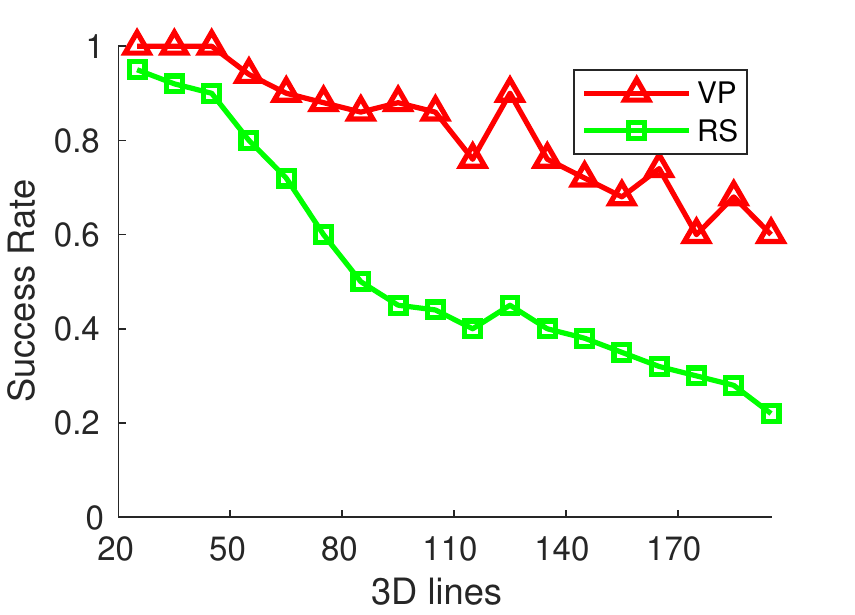} 
		\centerline{\scriptsize(\textbf{a}) Success rate} 
		\label{fig:sr}
	\end{minipage}
	\begin{minipage}[b]{0.32\linewidth}
		\centering
		\includegraphics[width = .9\columnwidth]{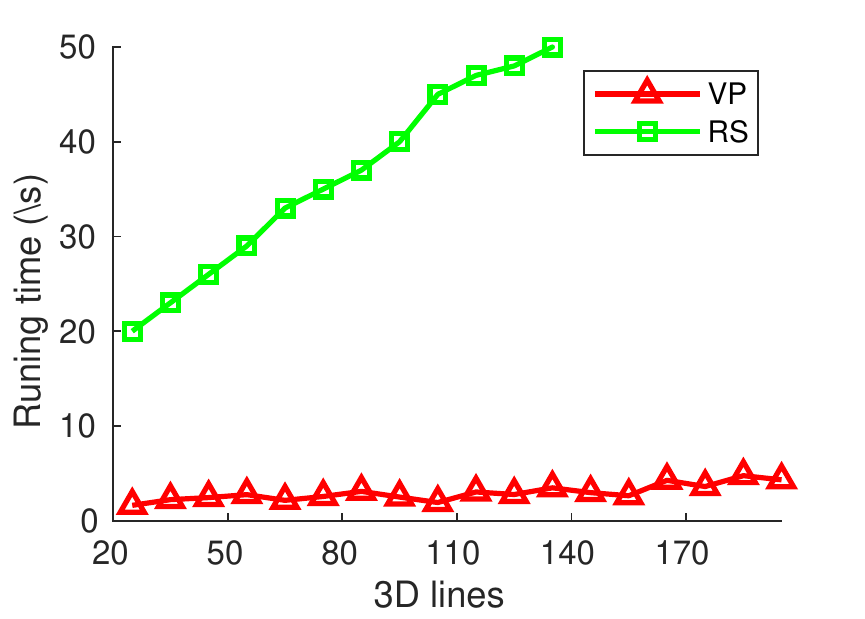}
		\centerline{\scriptsize(\textbf{b}) Runtime} 
		\label{fig:pose}
	\end{minipage}
	\begin{minipage}[b]{0.32\columnwidth}
		\centering
		\includegraphics[width = .99\columnwidth]{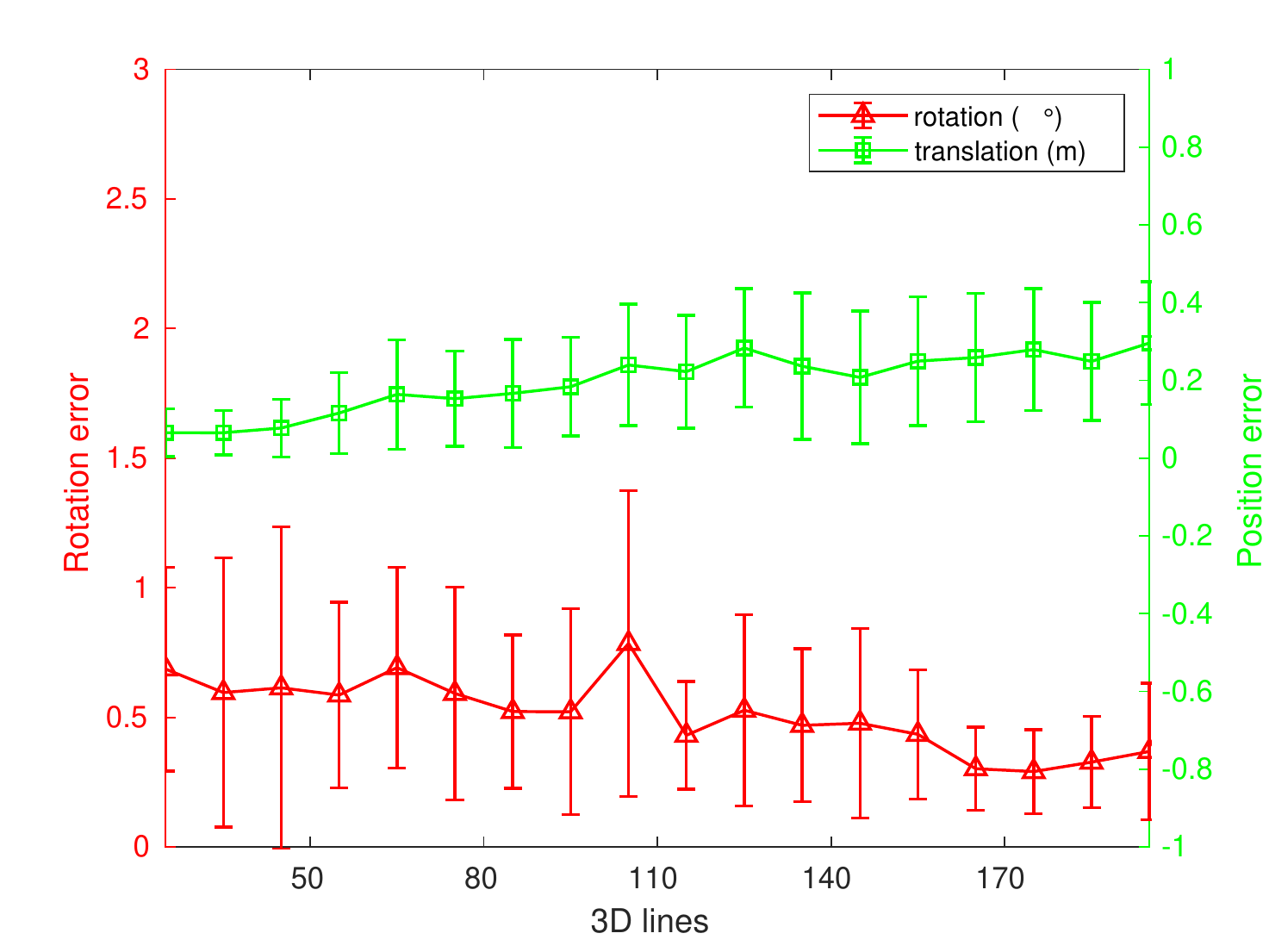}
		\centerline{\scriptsize(\textbf{c}) Camera pose error} 
		\label{fig:poseerror}
	\end{minipage}
	\caption{Results for synthetic dataset with different number of 3D lines. 50 Monte Carlo simulations are conducted for each setting.}
	\label{fig::analysis}
\end{figure}

\begin{figure}[hbtp]
	\centering
	\begin{minipage}[b]{0.40\columnwidth}
		\centering
		\includegraphics[width = .99\columnwidth]{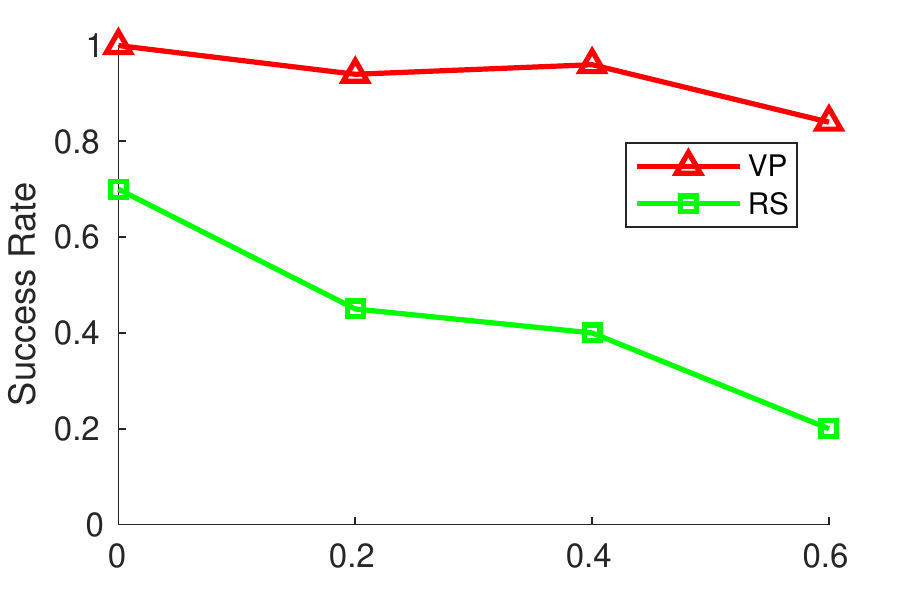}
		\centerline{\scriptsize(\textbf{a}) 3D outlier fraction} 
		\label{fig:3dout}
	\end{minipage}
	\begin{minipage}[b]{0.40\columnwidth}
		\centering
		\includegraphics[width = .99\columnwidth]{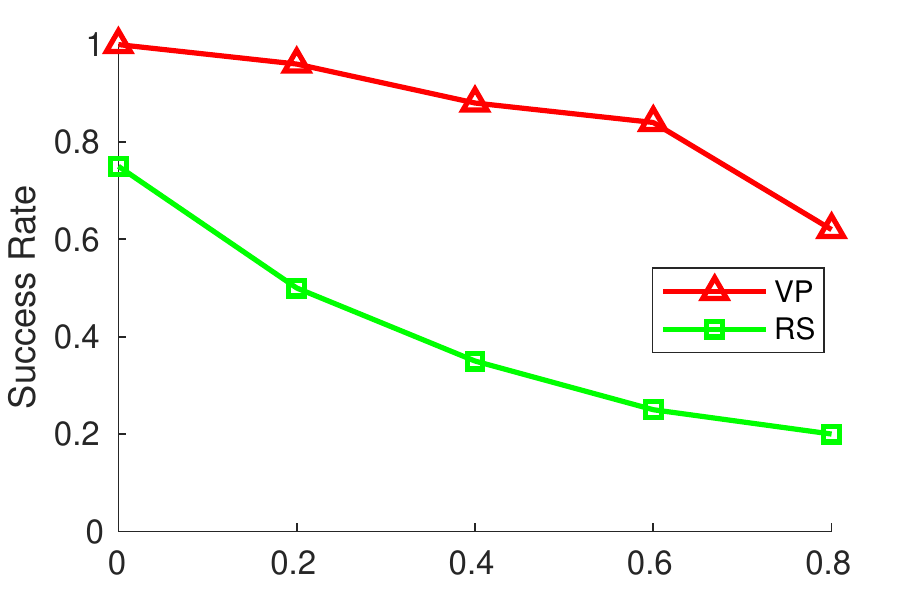}
		\centerline{\scriptsize(\textbf{b}) 2D outlier fraction} 
		\label{fig:2dout}
	\end{minipage}
	\caption{Outlier analysis. (a) Mean success rates with different 3D outlier fraction. (b) Mean success rates with different 2D outlier fraction.  }
	\label{fig:outlier}
\end{figure}

The quantitative results are shown in Fig.\ref{fig::analysis} and Fig.\ref{fig:outlier}. The success rates measure the fraction of trials where the correct pose is found, where the rotation error $R_e=acos(\frac{trace(R_a^TR_b)-1}{2})$ is less than 0.1 radians and position error related to the ground truth $||t_e||/||t_{GT}||$ is less than 0.1, as used in \cite{campbell2018globally}. Compared with RANSAC (RS), the proposed method (VP) has a higher success rate with the growth of line feature numbers. {\color{black}When there are more lines, there will be more number of outliers. In this case, it will be easier to fall into a local optimum, thus the successful pose estimation rate will drop.} The running time becomes a little longer but the growth rate is much smaller than RANSAC. {\color{black}The time costs are less than 5 seconds for each trail. Fig.\ref{fig::analysis}(c) shows the RMSE and the standard deviations of the pose estimation errors. The rotation error is less than 1.5 degrees, while the position error is less than 0.4 meters.}  Additionally, we can observe from Fig.\ref{fig:outlier} that the proposed method is very robust to 2D and 3D outliers.

\subsection{Real data experiments.} \label{real}
 The dataset consists of four outdoor and indoor scenes of structured environment, CMU NSH wall, Hamburg Hall windows, Hamerschlag Hall and NSH lounge. For each scene, there are a point cloud and 10 images taken with different poses. A FARO laser scanner focus3D S and FLIR BFLY-U3-13S2C-CS camera are used to capture 3D point clouds and 2D images. {\color{black}{To get the ground truth of each camera pose, we manually and uniformly pick 20 pairs of 2D-3D correspondences for each image, then use the PnP solver to find camera pose with minimal projection error. It is time-consuming but we trust the manually labeled 2D-3D correspondences.}} %It should be noticed that the captured point clouds have color information, which is only used to obtain the true camera poses. Without the color information, it is difficult to obtain the ground truth of camera poses. Because image-to-image registration can be solved very well by point feature matching, we guess a rough camera pose and project the colored point cloud to form a colored image. Then we can obtain 2D-3D point correspondence by matching SIFT features of the projected colored image with camera image. Based on the 2D-3D point correspondence and the pre-calibrated intrinsic camera matrix $K$, typical PnP solver is used to calculate the camera pose. We set this computed pose as the ground truth for analysis. %The details of the dataset will be available online soon. %\footnote{https://github.com/levenberg/2D-3D-registration-dataset.git}.
%  \begin{figure}[h]
% 	\begin{center}
% 		\includegraphics[width=0.6\linewidth]{sensorsetup_compressed.pdf}
% 	\end{center}
% 	\caption{Sensor setup for collecting 2D and 3D data.}
% 	\label{fig:sensor}
% \end{figure} 

\begin{figure}[t]
 	\centering
 	\begin{minipage}[b]{0.49\linewidth}
 		\centering
 		\includegraphics[width = .99\columnwidth]{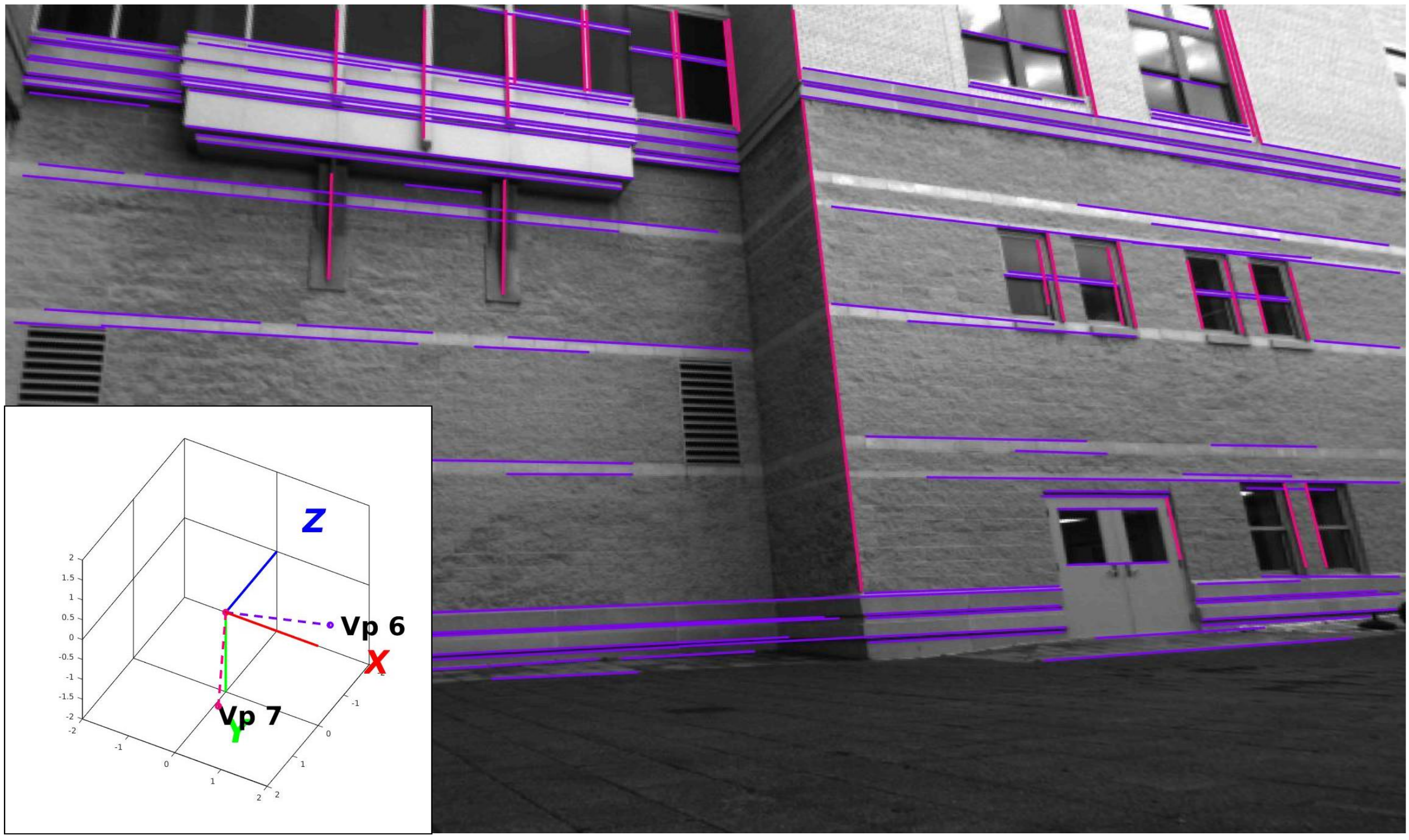} 
 		%\centerline{\scriptsize(\textbf{a}) Vanishing directions and associated 2D lines } 
 		\label{fig:2dvp}
 	\end{minipage}
 	\begin{minipage}[b]{0.49\linewidth}
 		\centering
 		\includegraphics[width = .99\columnwidth]{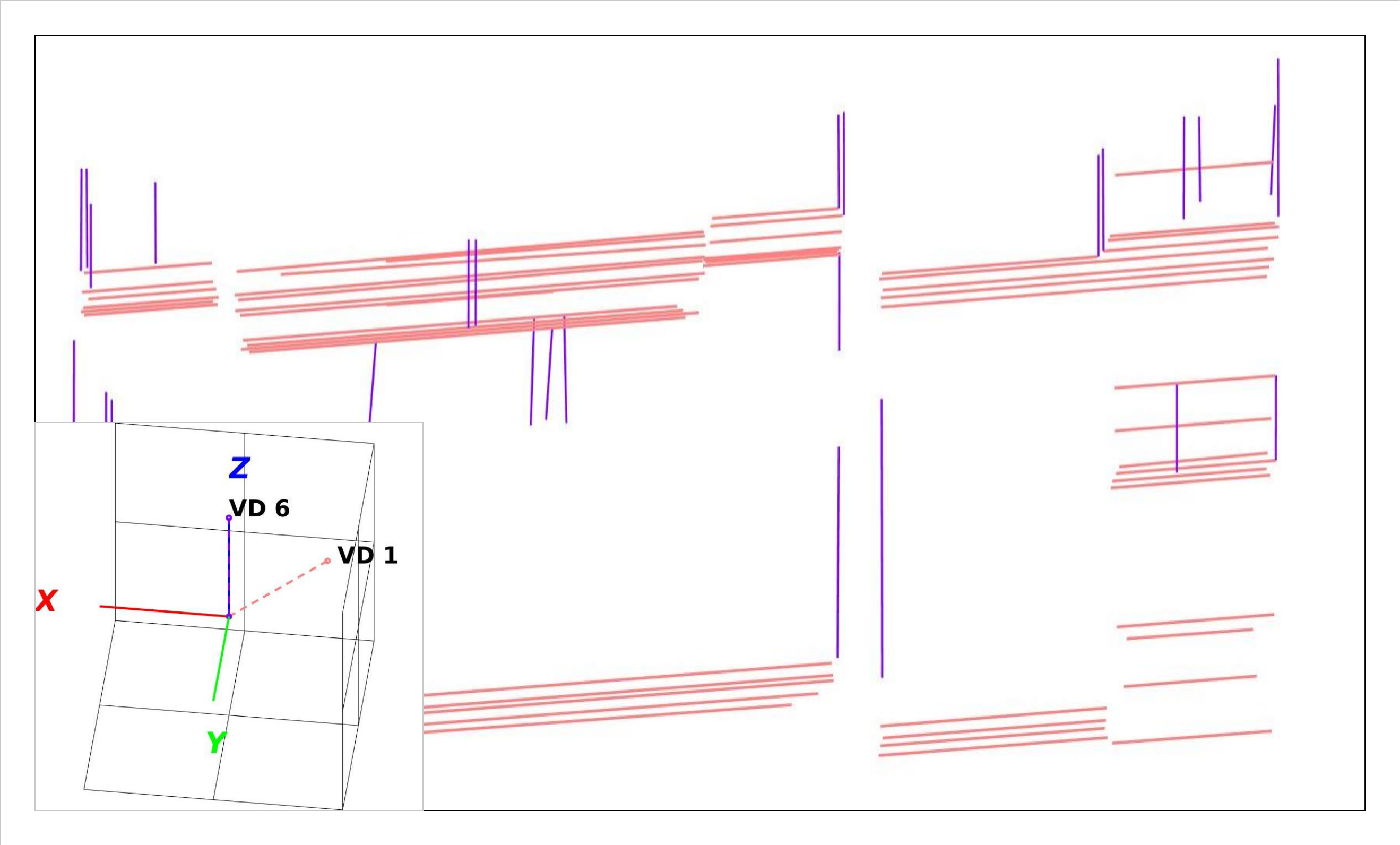}
 		%\centerline{\scriptsize(\textbf{b}) 3D orientations and associated 3D lines} 
 		\label{fig:3dvd}
 	\end{minipage}
 	\caption{Vanishing direction to 3D orientation correspondence. Left: Vanishing directions and associated 2D lines, right: 3D orientations and associated 3D lines  }
 	\label{fig:vpvd}
 \end{figure}

To validate the effectiveness of the rotation estimation framework, we show an example of the estimated vanishing directions and the corresponding 2D line segment sets in Fig.\ref{fig:vpvd}(a), while the primary 3D orientations and the corresponding 3D parallel line segments in Fig.\ref{fig:vpvd}(b). The two primary vanishing directions for 2D image are corresponding to the two primary 3D line orientations. However, without visualization of 2D lines in images and 3D lines in point clouds, it is difficult to compute 2D-3D direction correspondence. Thus, we keep the possibilities of orientation matching and get 8 rotation matrix candidates. % and pick the one with the most number of line correspondences in the translation estimation procedure. %The number of line segments is also reduced by merging in each set, which will greatly increase the efficiency of later translation estimation. 

\begin{figure}[t]
	\centering
	\begin{minipage}[b]{1.1\linewidth}
		\centering
		\includegraphics[width = .32\columnwidth]{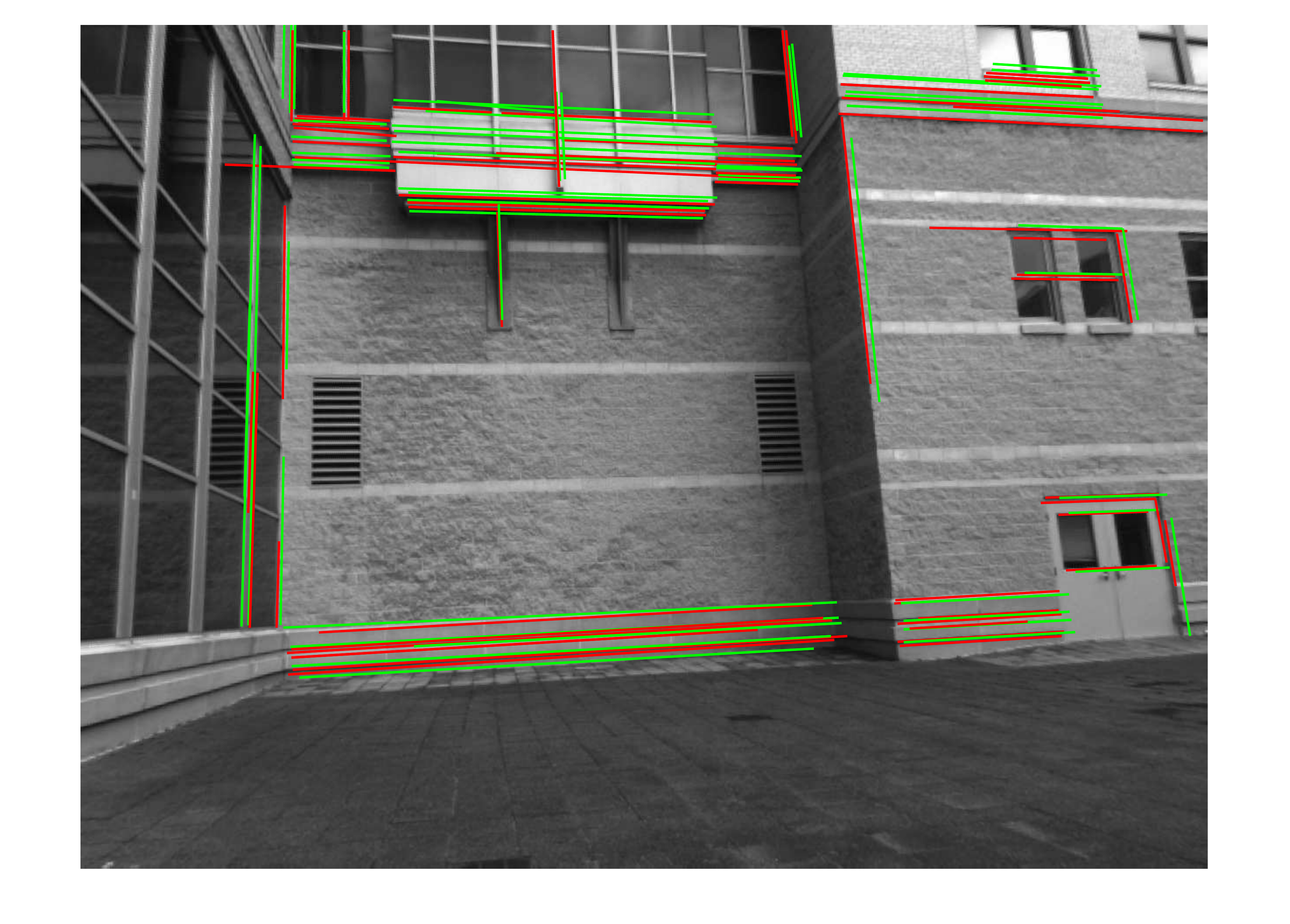}
		\includegraphics[width = .32\columnwidth]{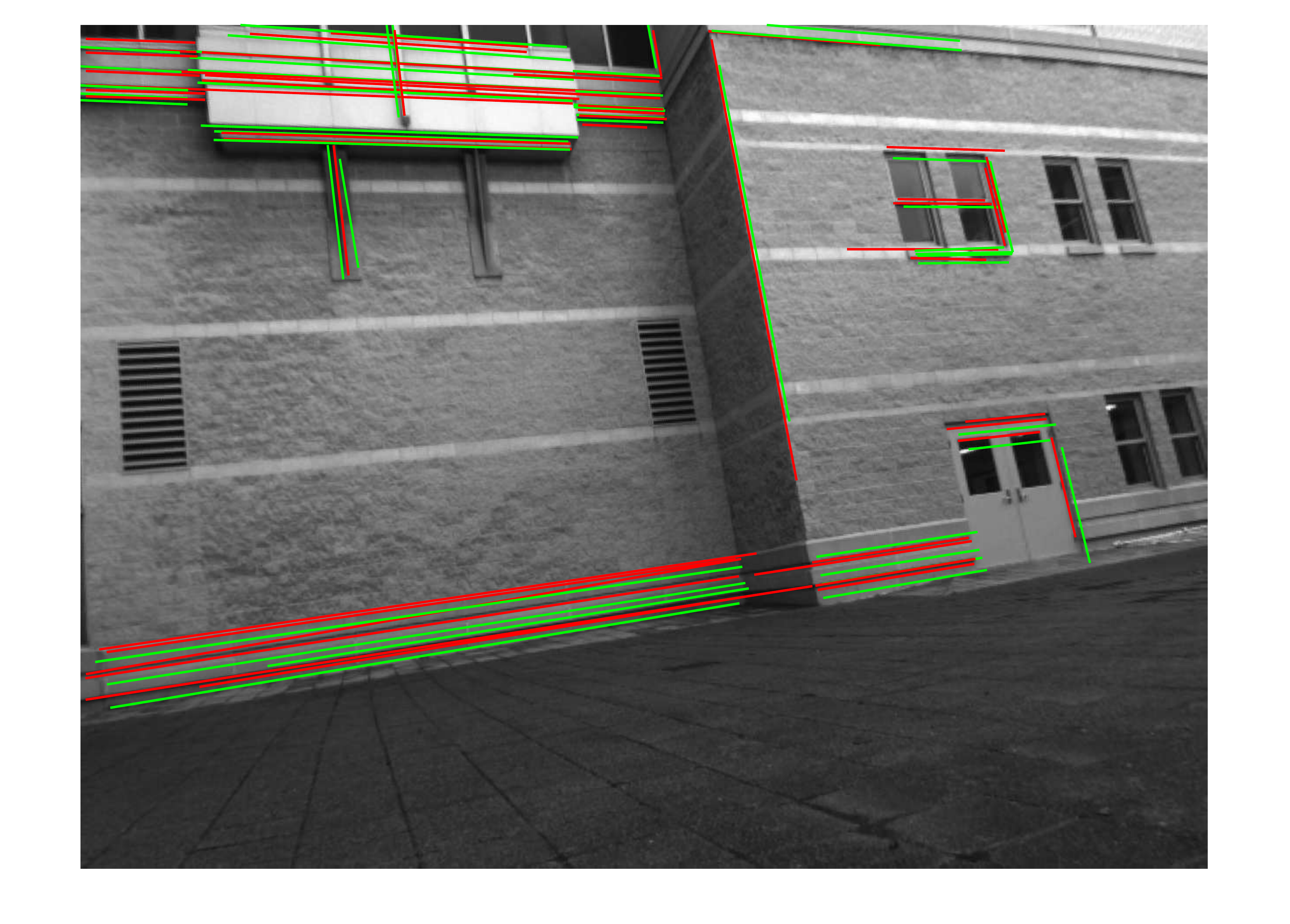}  
		\includegraphics[width = .32\columnwidth]{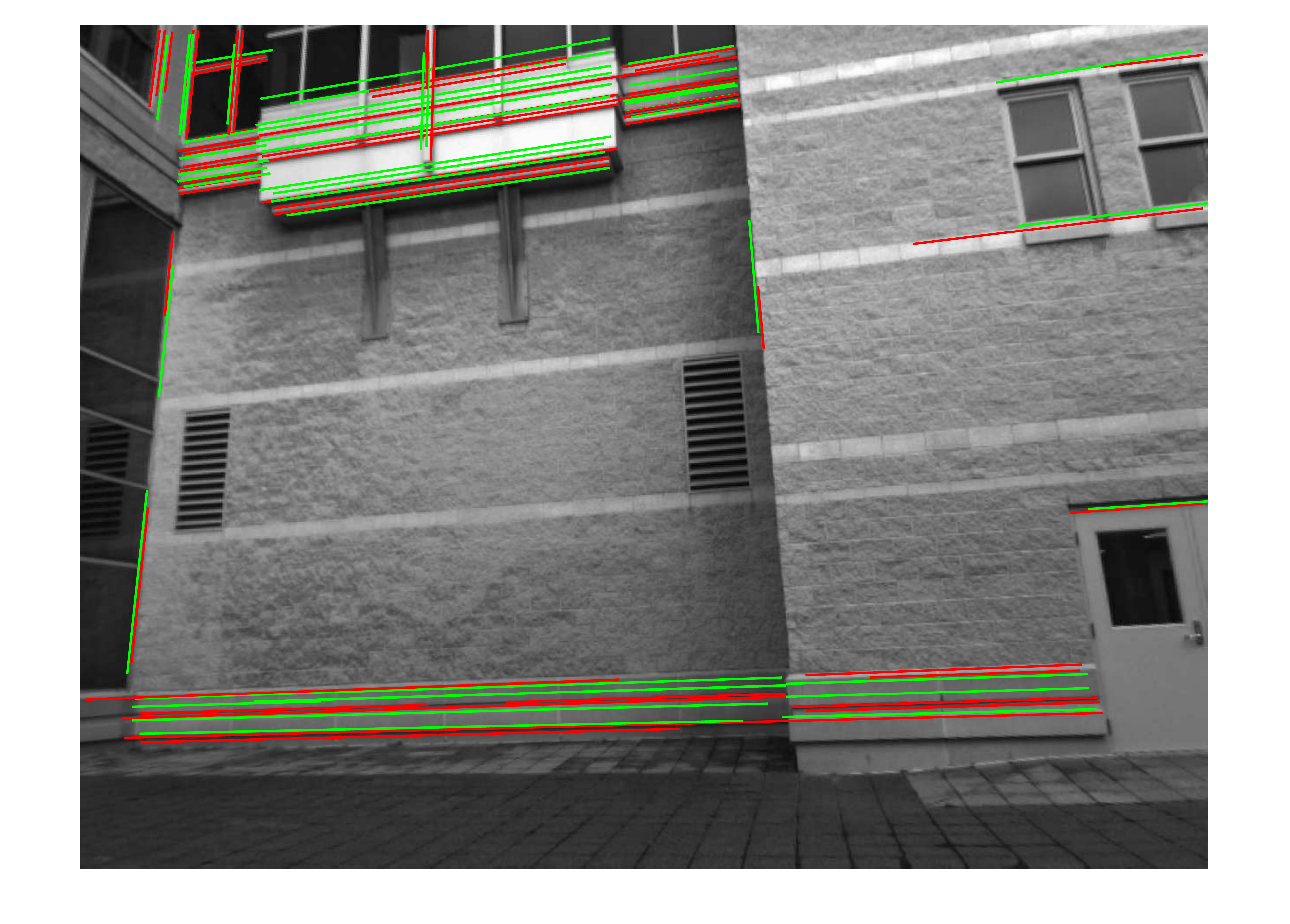}  
		\centerline{\scriptsize(\textbf{a}) NSH wall. } 
		\label{fig:scene1}
	\end{minipage}
	\begin{minipage}[b]{1.1\linewidth}
		\centering
		\includegraphics[width = .32\columnwidth]{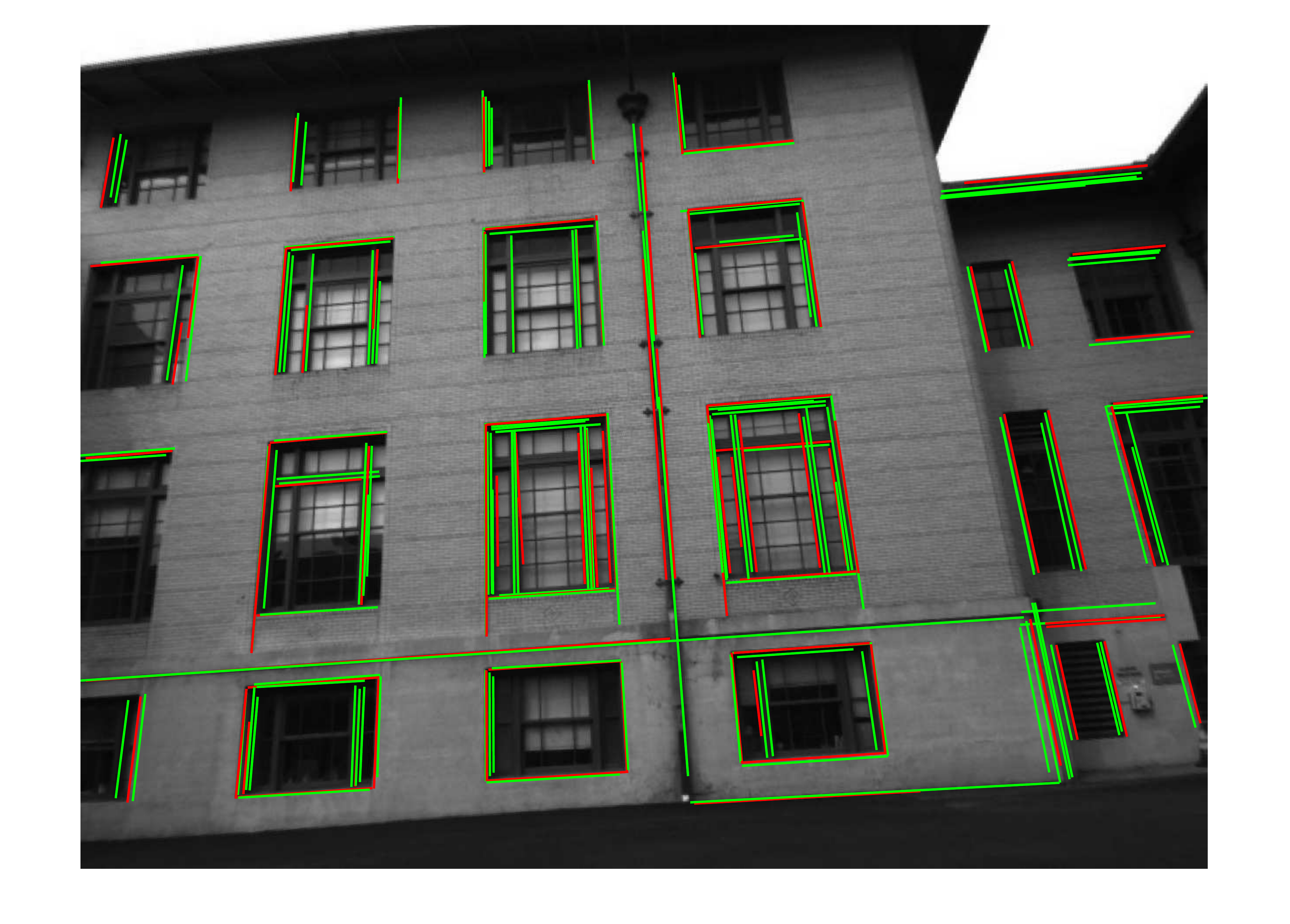}
		\includegraphics[width = .32\columnwidth]{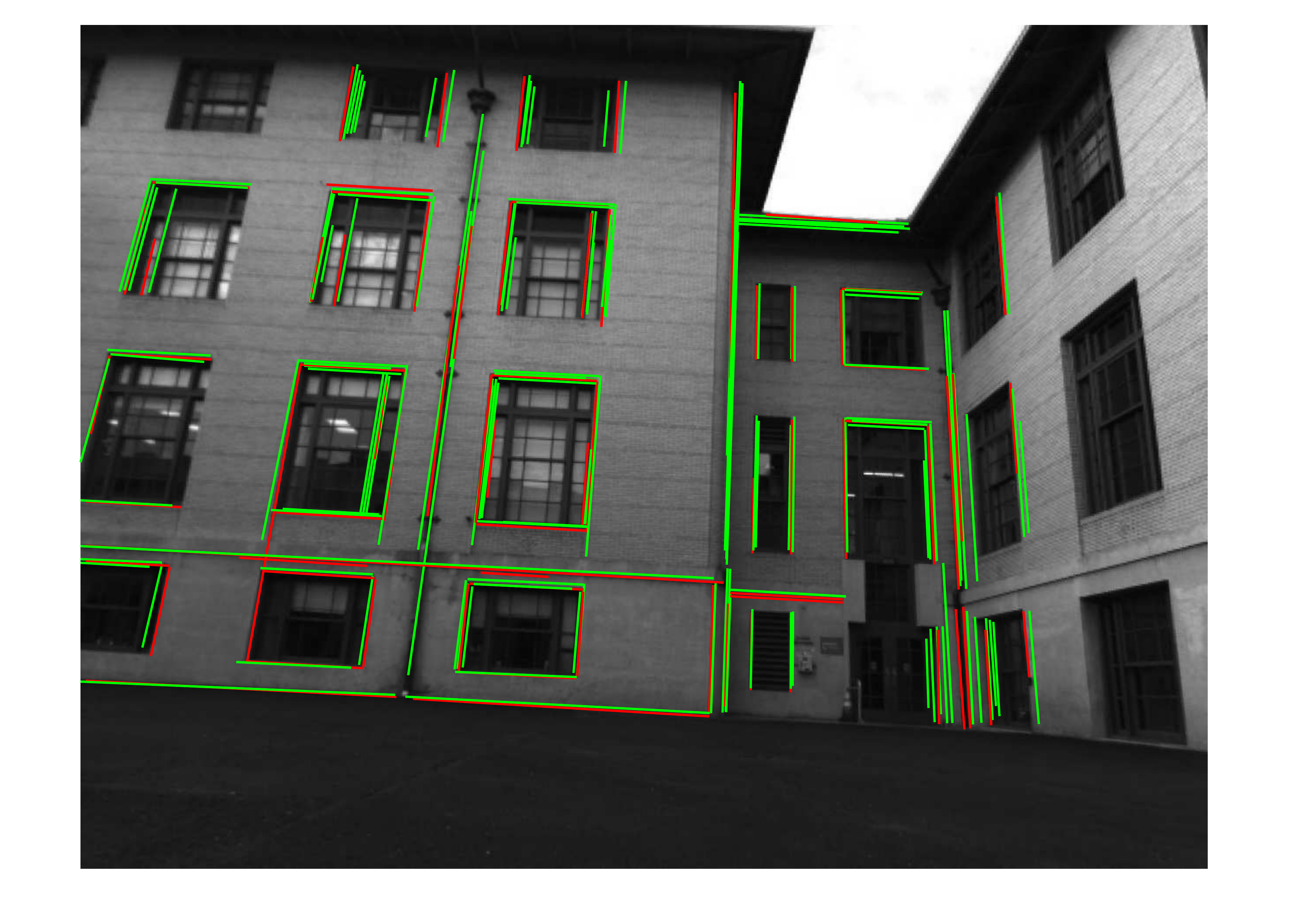}  
		\includegraphics[width = .32\columnwidth]{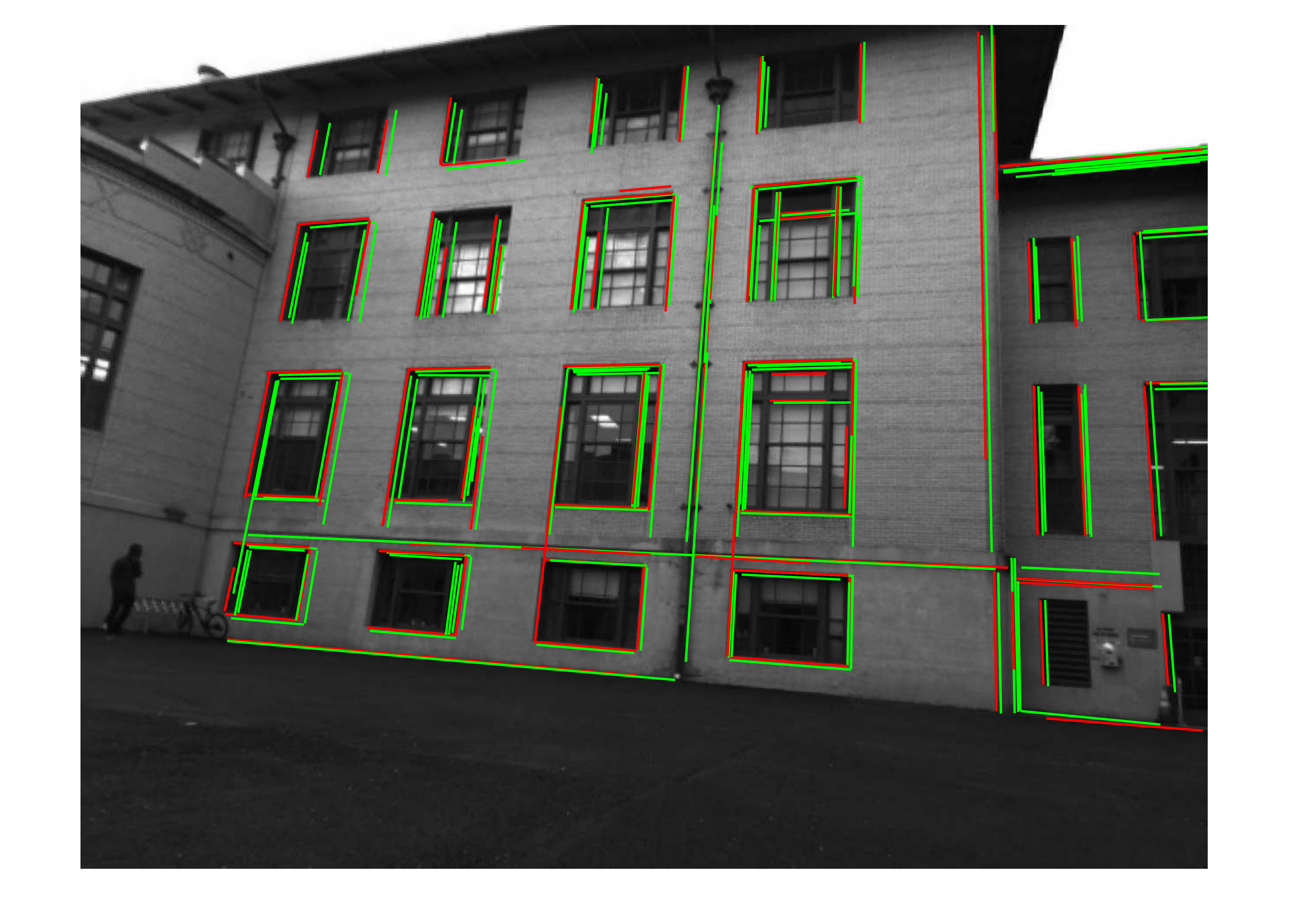}  
		\centerline{\scriptsize(\textbf{b}) Hamburg Hall windows.} 
		\label{fig:scene2}
	\end{minipage}
\begin{minipage}[b]{1.1\linewidth}
	\centering
	\includegraphics[width = .32\columnwidth]{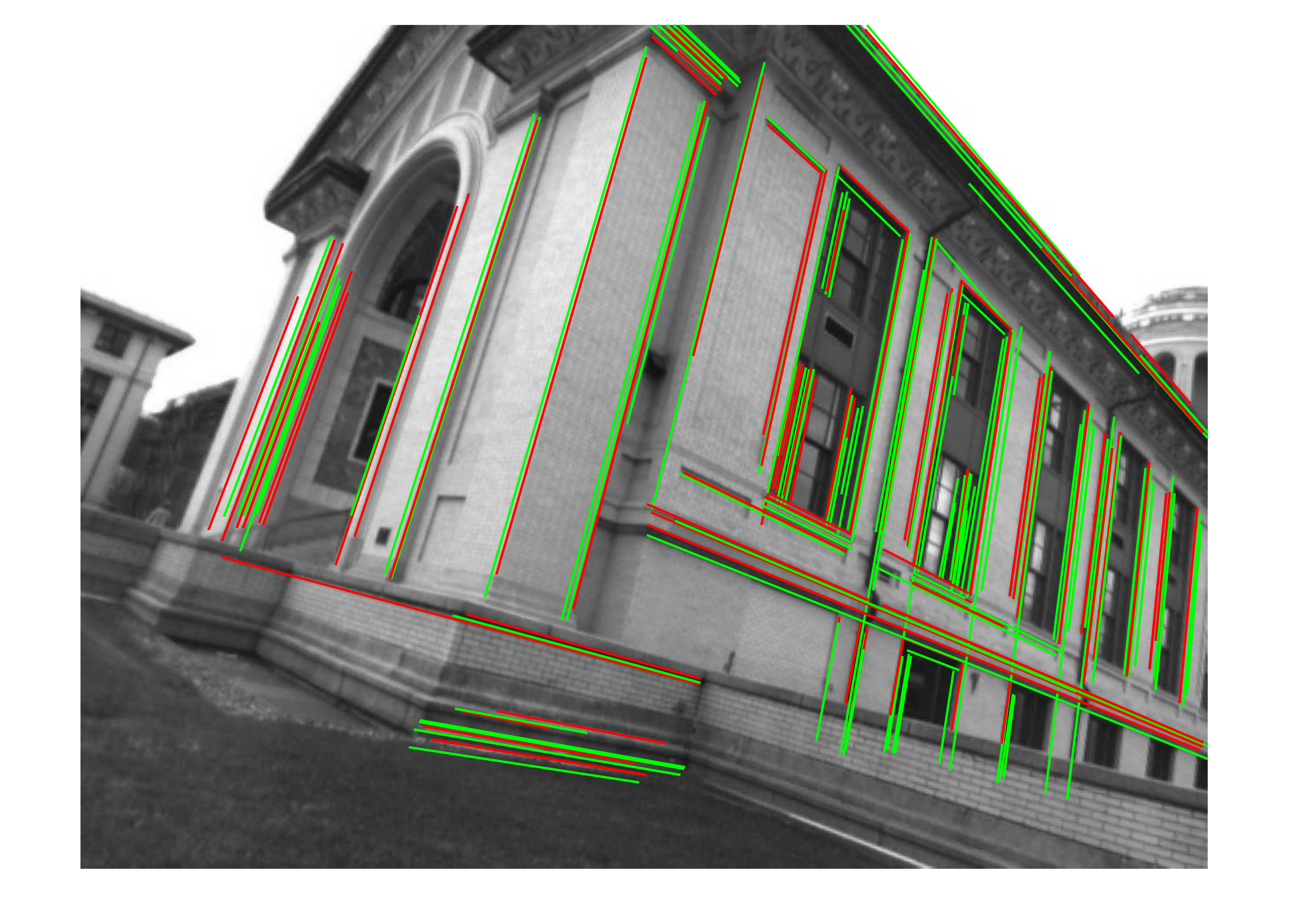}
	\includegraphics[width = .32\columnwidth]{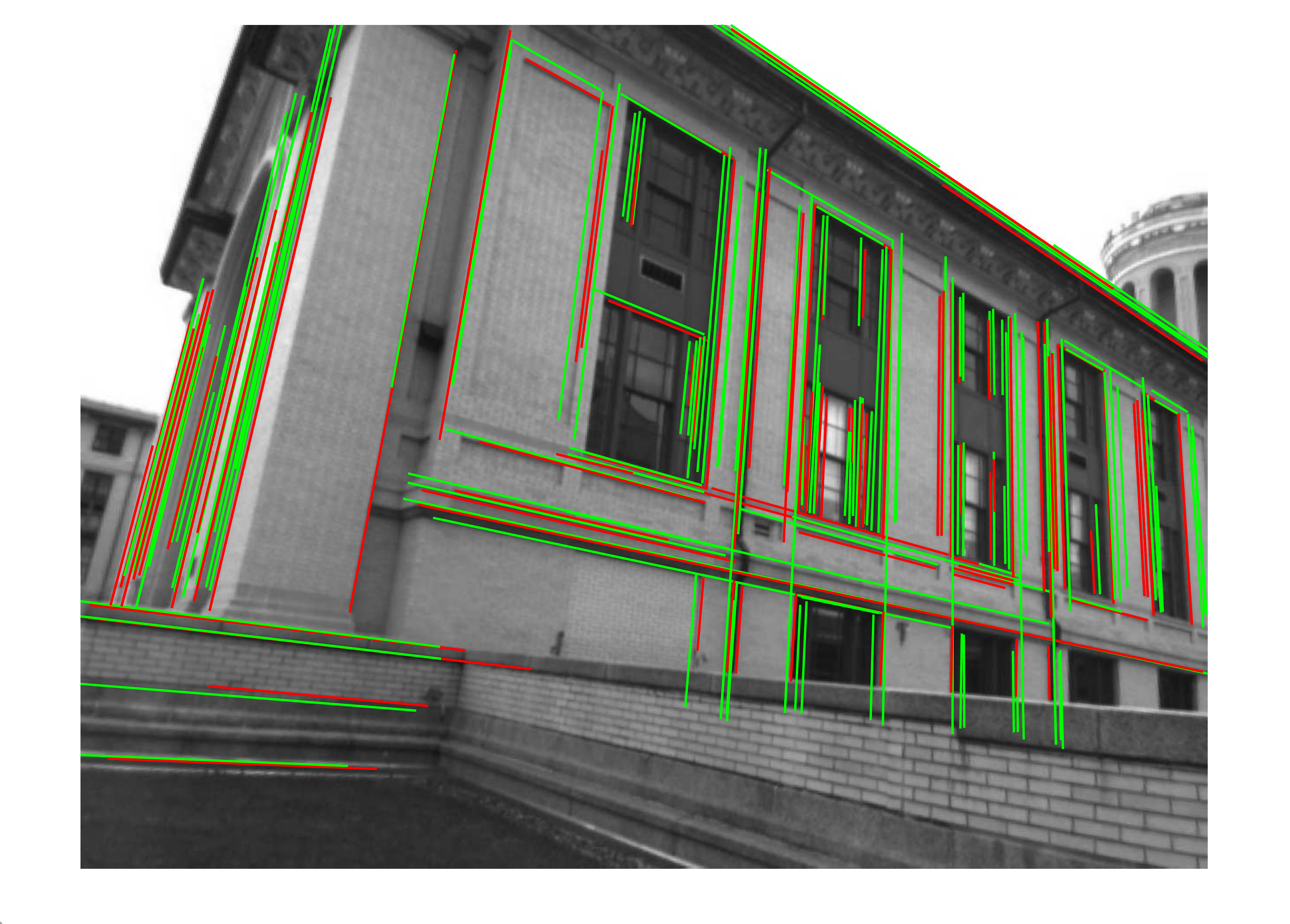}  
	\includegraphics[width = .32\columnwidth]{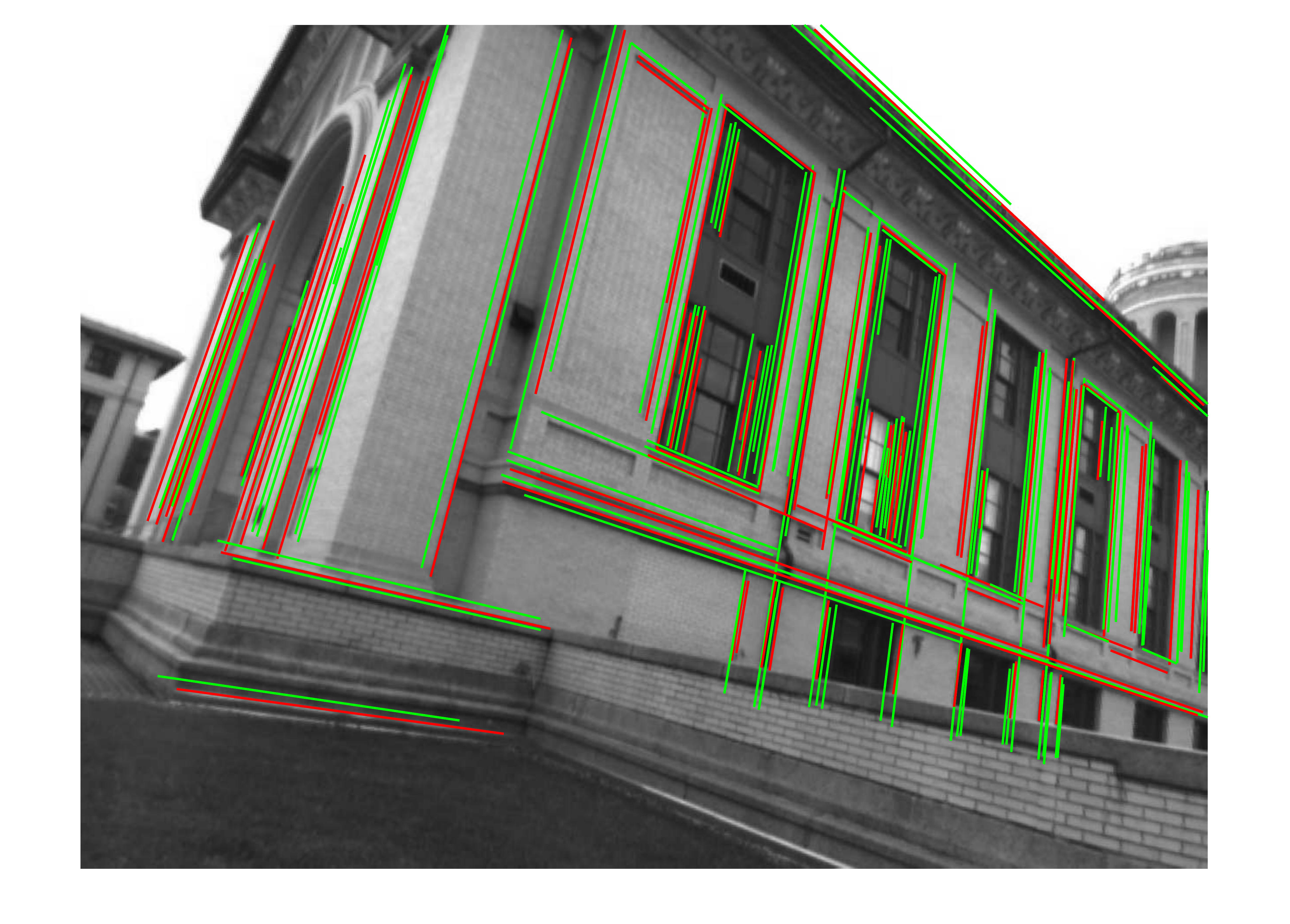}  
	\centerline{\scriptsize(\textbf{c}) Hamerschlag Hall.} 
	\label{fig:scene3}
\end{minipage}
\begin{minipage}[b]{1.1\linewidth}
	\centering
	\includegraphics[width = .32\columnwidth]{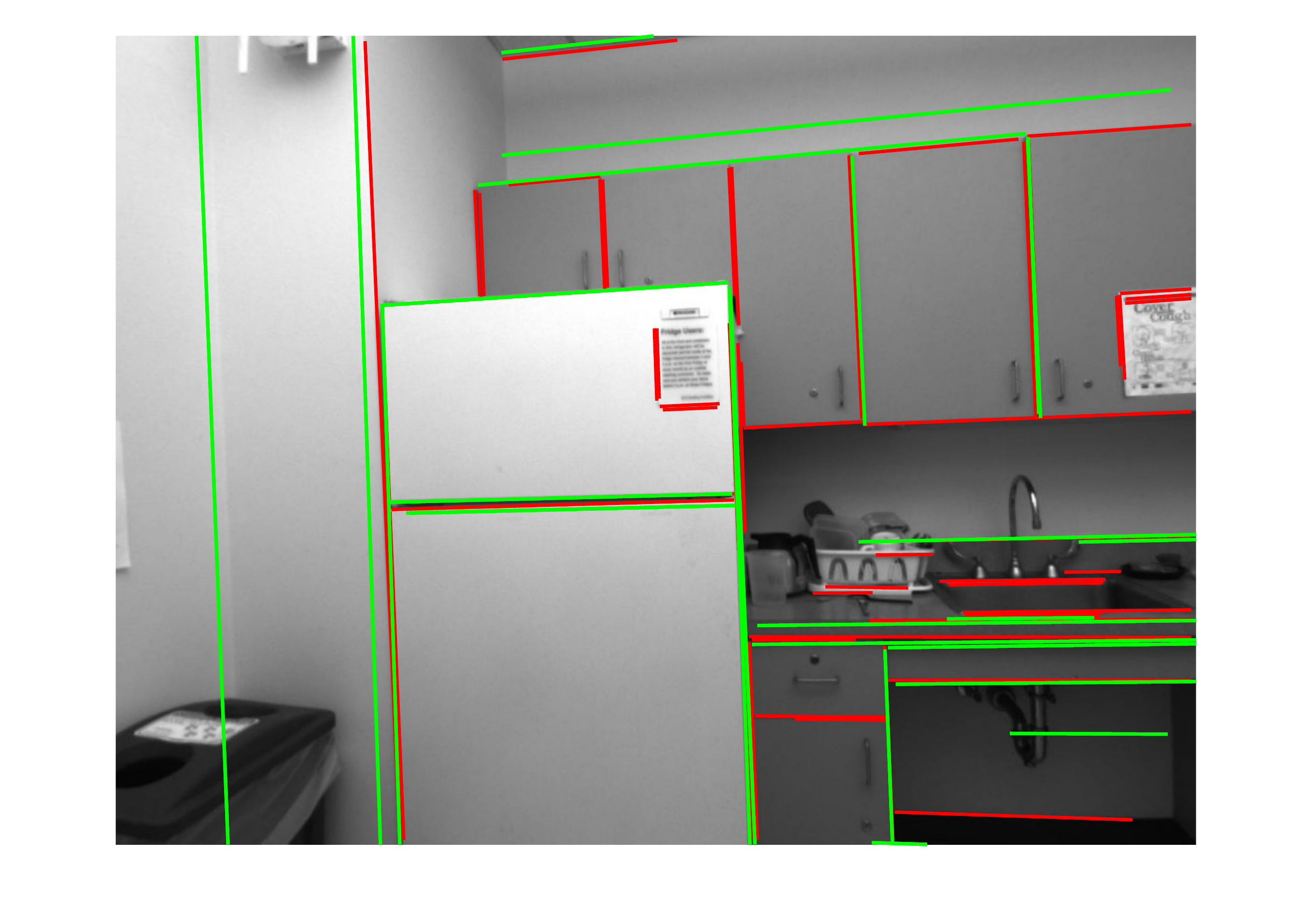}
	\includegraphics[width = .32\columnwidth]{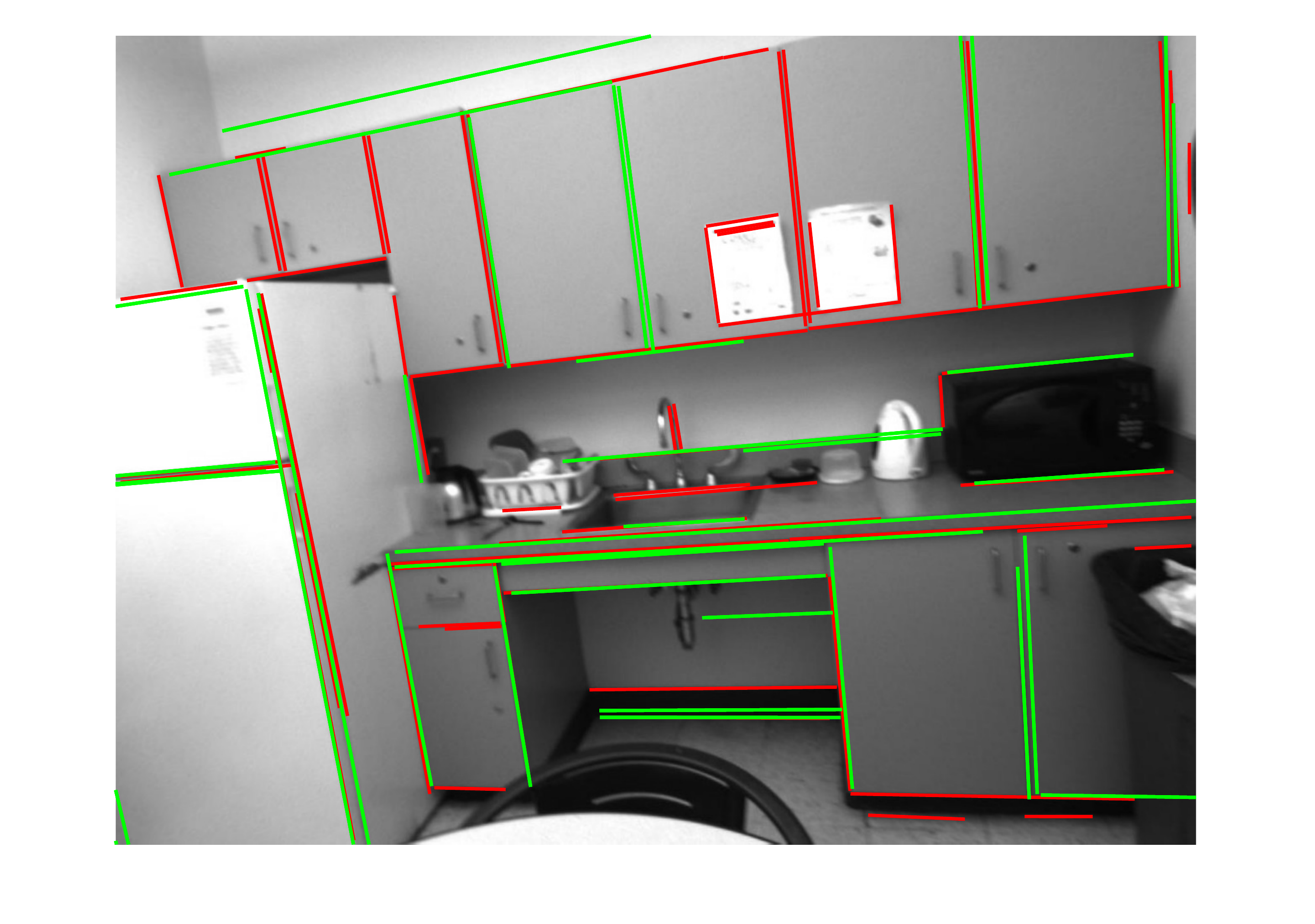}  
	\includegraphics[width = .32\columnwidth]{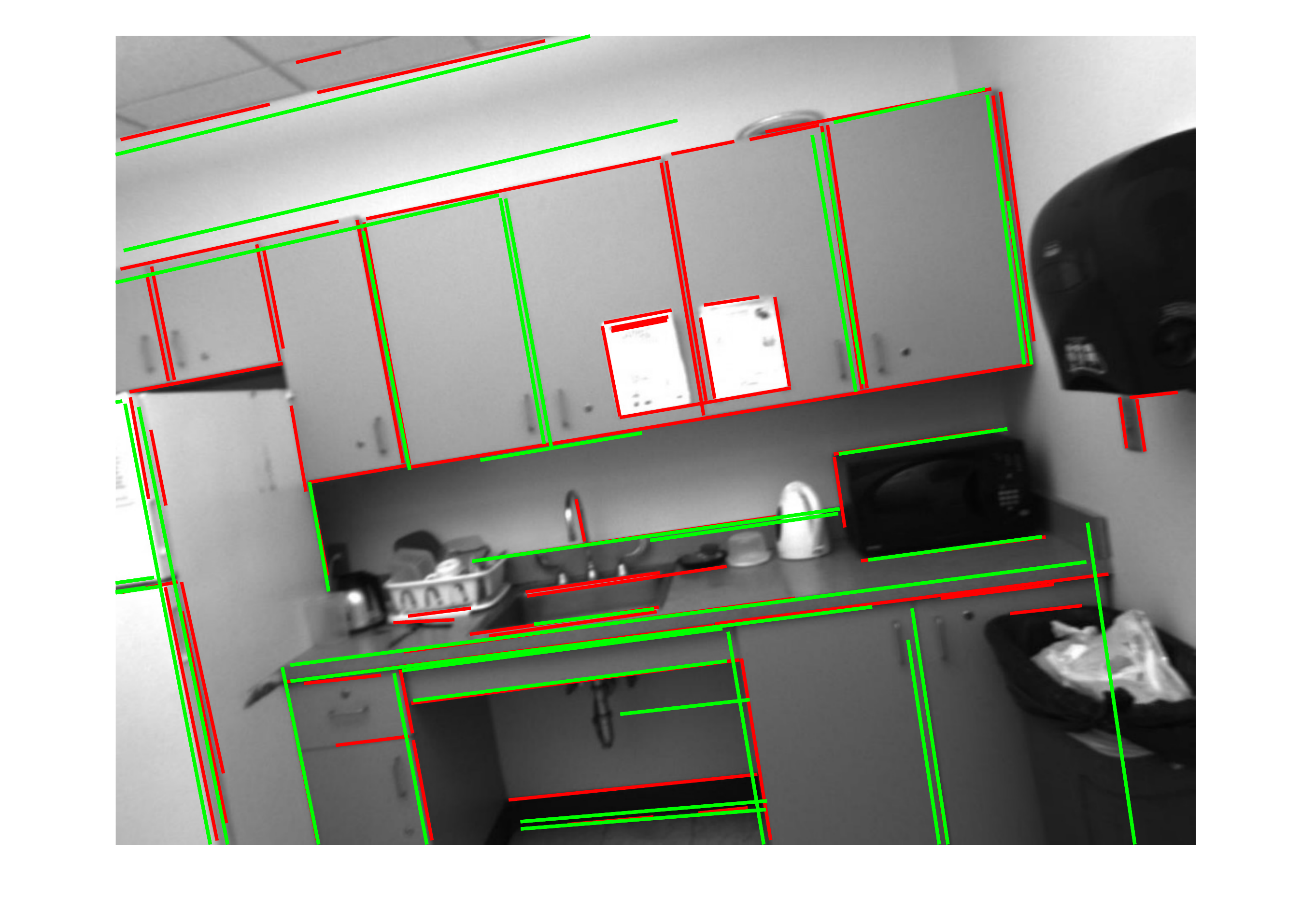}  
	\centerline{\scriptsize(\textbf{d}) NSH lounge.} 
	\label{fig:scene4}
\end{minipage}
	\caption{Demonstration of the 2D-3D line correspondence for three scenes. (Green: 3D line projections, red: 2D lines.) }
	\label{fig:linematch}
\end{figure}

Some qualitative results of line correspondences are shown in Fig.\ref{fig:linematch} for four scenes. All the matching 3D line segments are projected to the image plane (in green) using the estimated pose for visualization, while the red are the corresponding 2D lines. We can observe that the global geometric structure aligns well. Some 2D lines have more than one corresponding 3D lines and vice versa. This is reasonable because we can not guarantee that the fragments in 2D and 3D line segments are totally removed. There exists some (both 2D and 3D) lines having no correspondence because they contribute to the vanishing direction matching but not for translation estimation.  Based on these 2D-3D line correspondences and the coarse estimated pose, we further use the local optimization method in Sec.\ref{localrefine} to refine the pose. The iterations of pose refinement often less than 5 times when there is no outlier.

\begin{figure}[htbp]
	\centering
	\begin{minipage}[b]{0.49\linewidth}
		\centering
		\includegraphics[width = .99\columnwidth]{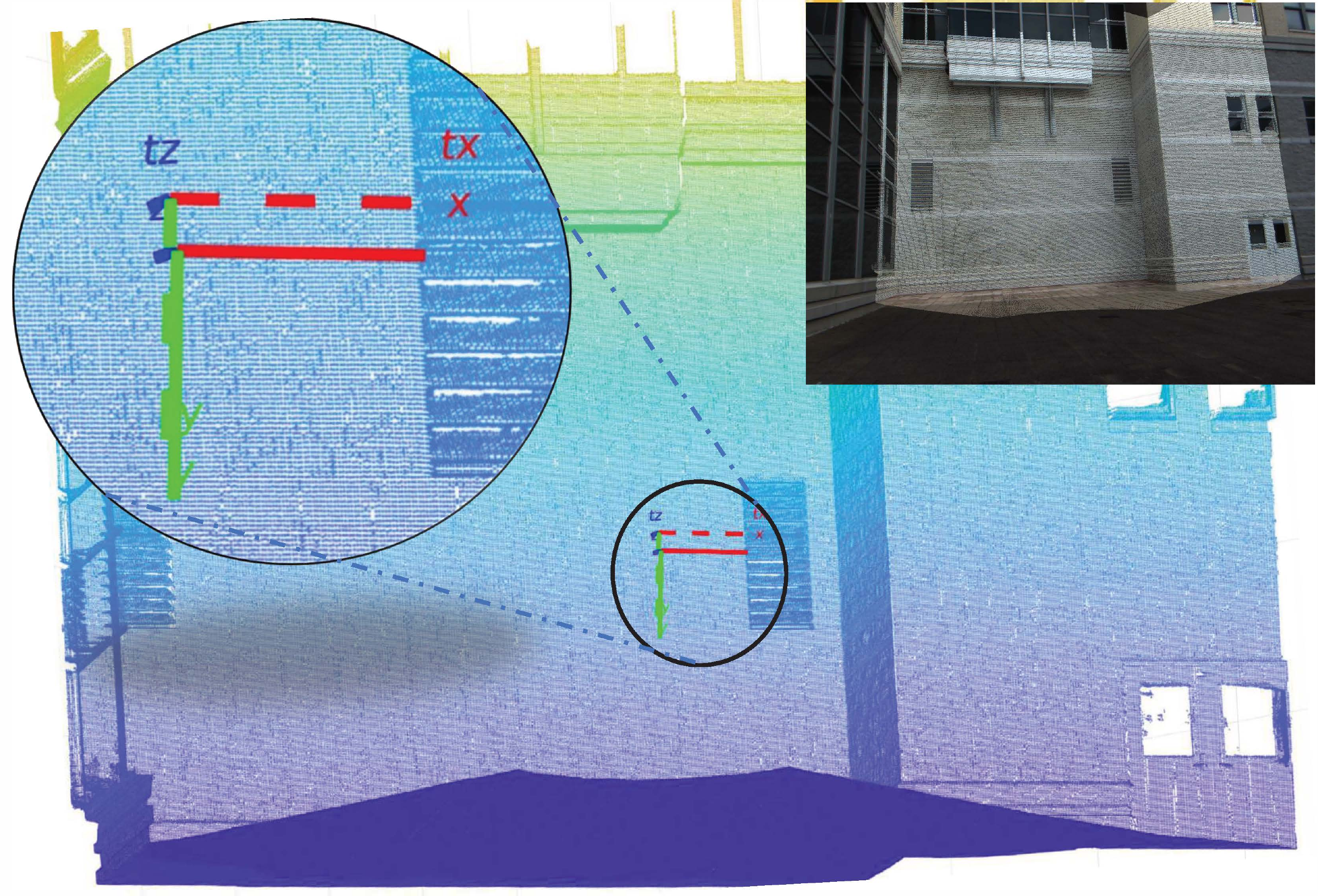} 
		%\centerline{\scriptsize(\textbf{a}) camera pose. } 
		\label{fig:cm1}
	\end{minipage}
% 	\begin{minipage}[b]{0.49\linewidth}
% 		\centering
% 		\includegraphics[width = .90\columnwidth]{overlap1.jpg}
% 		%\centerline{\scriptsize(\textbf{b}) point cloud projection.} 
% 		\label{fig:pcp1}
% 	\end{minipage}
	\begin{minipage}[b]{0.49\linewidth}
		\centering
		\includegraphics[width = .99\columnwidth]{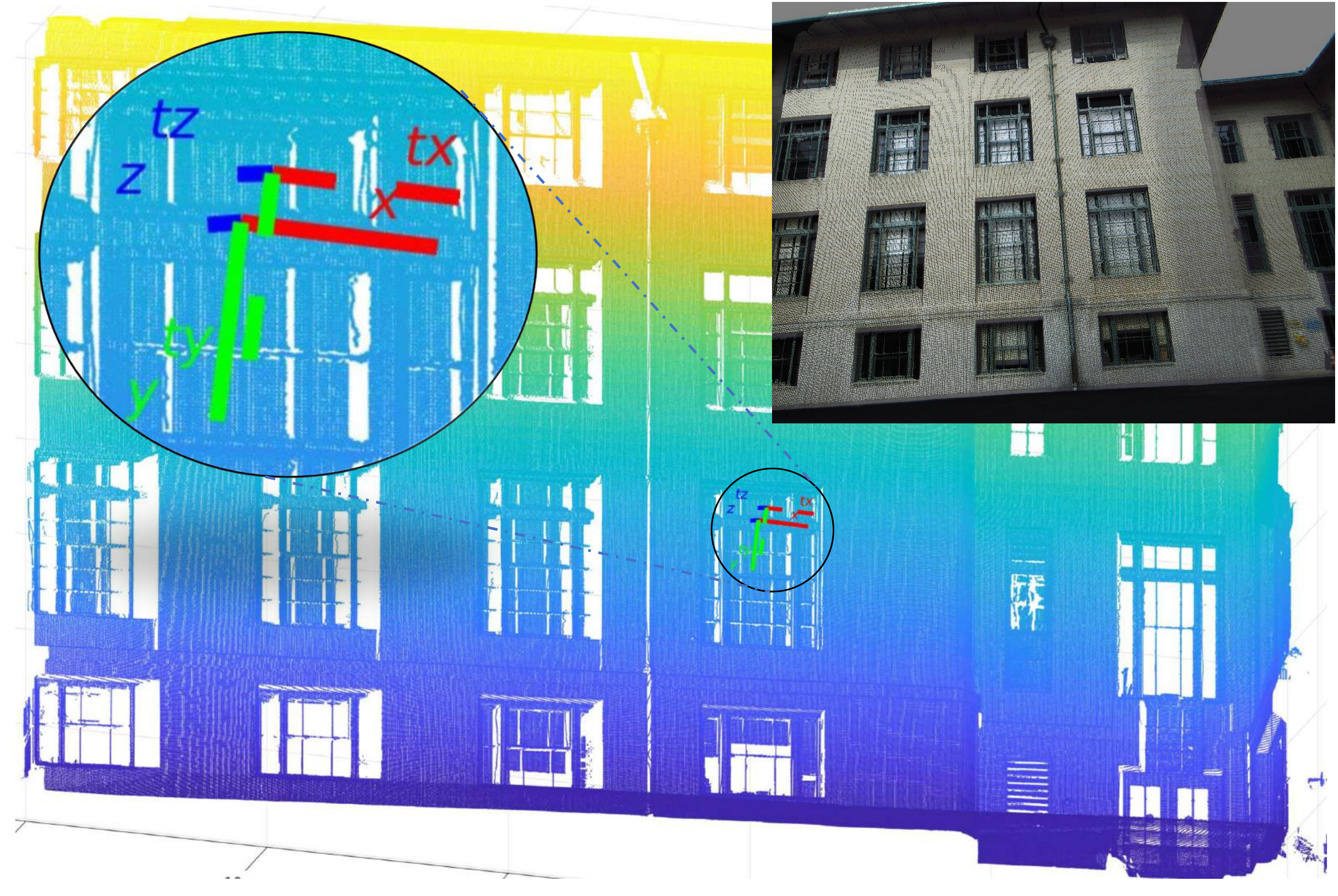} 
		%\centerline{\scriptsize(\textbf{a}) camera pose. } 
		\label{fig:cm2}
	\end{minipage}
% 	\begin{minipage}[b]{0.49\linewidth}
% 		\centering
% 		\includegraphics[width = .90\columnwidth]{overlap2.jpg}
% 		%\centerline{\scriptsize(\textbf{b}) point cloud projection.} 
% 		\label{fig:pcp2}
% 	\end{minipage}
	\begin{minipage}[b]{0.49\linewidth}
		\centering
		\includegraphics[width = .99\columnwidth]{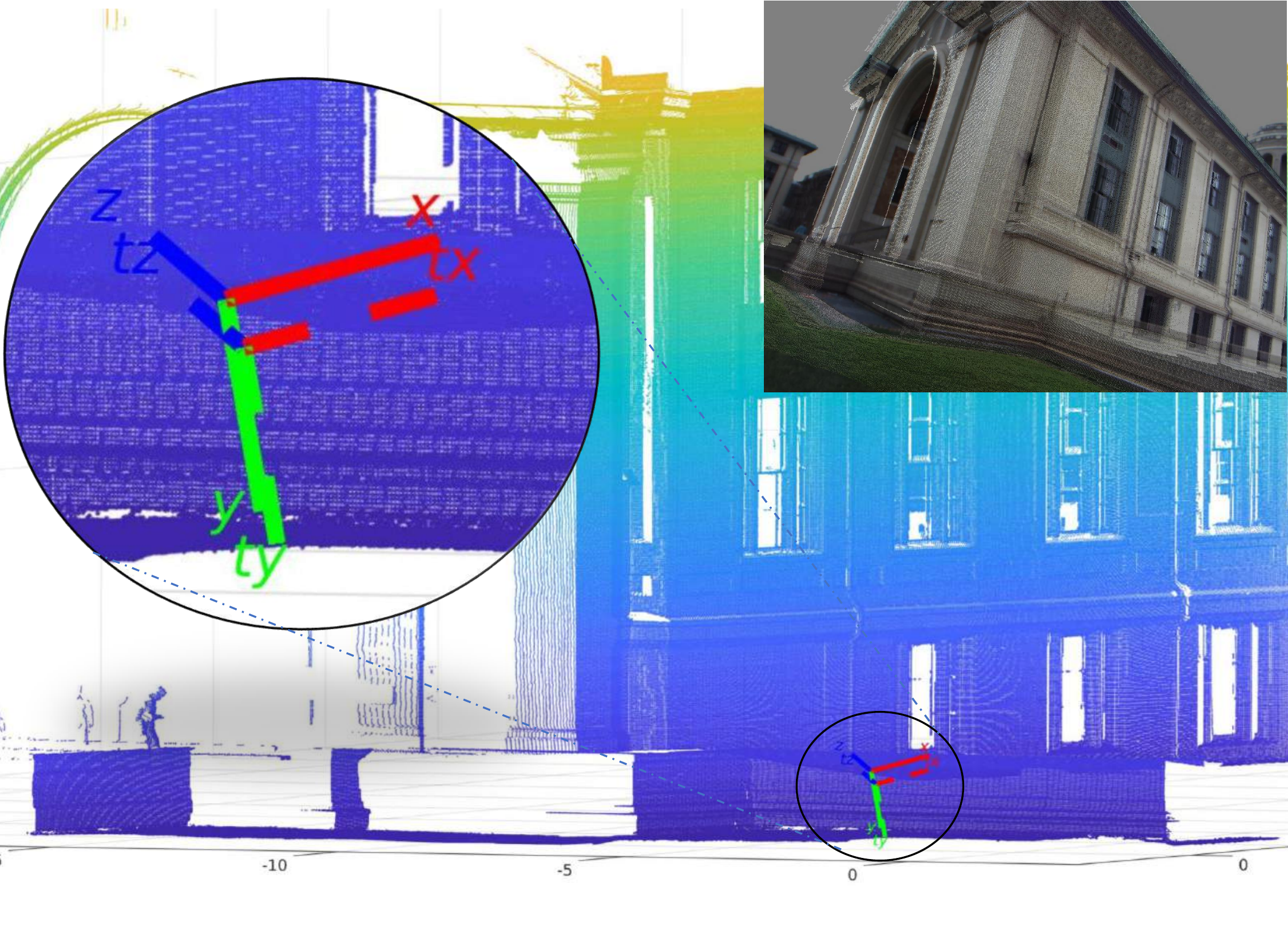} 
		%\centerline{\scriptsize(\textbf{a}) camera pose. } 
		\label{fig:cm3}
	\end{minipage}
% 	\begin{minipage}[b]{0.49\linewidth}
% 		\centering
% 		\includegraphics[width = .90\columnwidth]{overlap3.jpg}
% 		%\centerline{\scriptsize(\textbf{b}) point cloud projection.} 
% 		\label{fig:pcp3}
% 	\end{minipage}
	\begin{minipage}[b]{0.49\linewidth}
		\centering
		\includegraphics[width = .99\columnwidth]{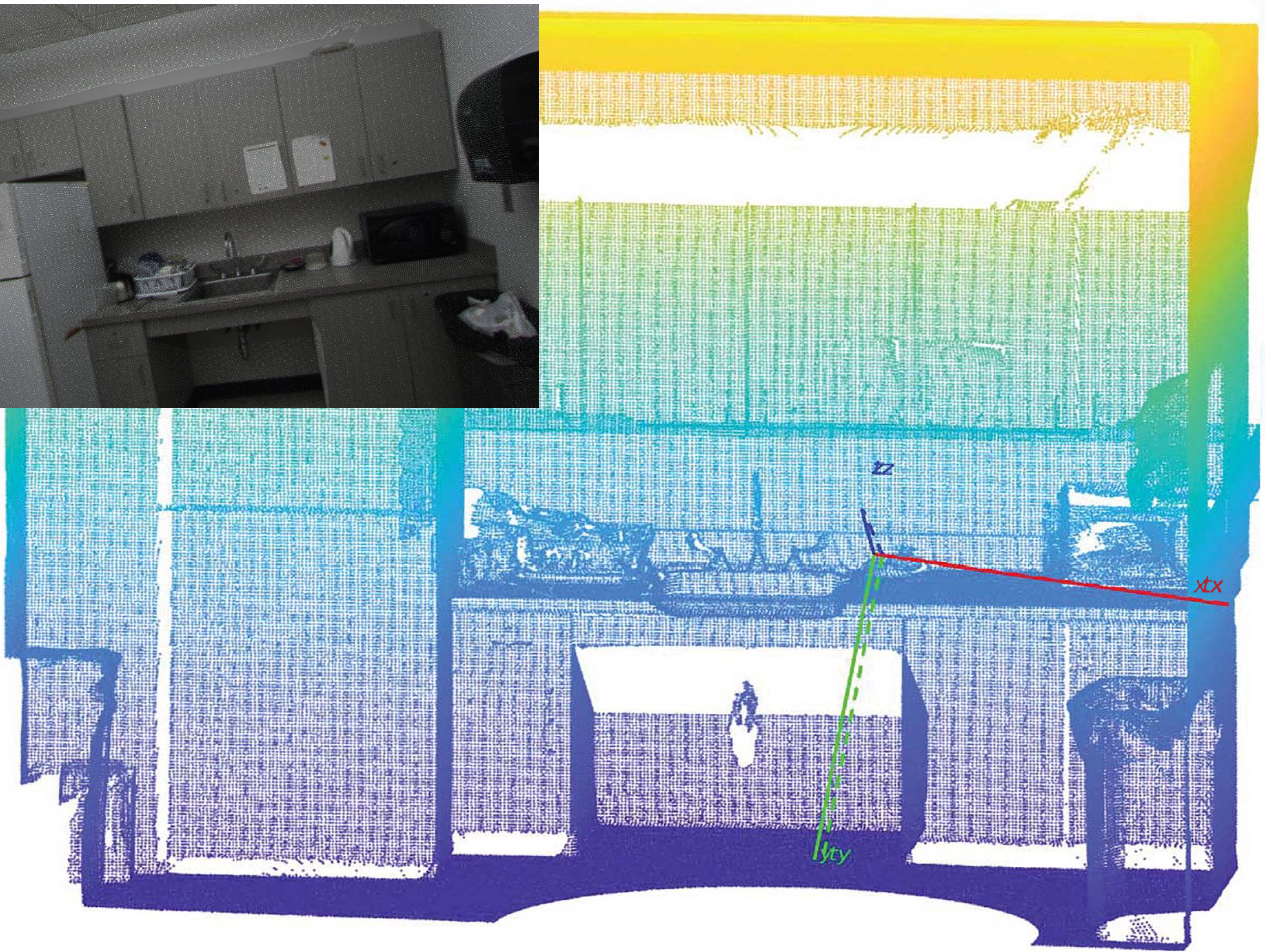} 
		%\centerline{\scriptsize(\textbf{a}) camera pose. } 
	    \label{fig:cm4}
	\end{minipage}
% 	\begin{minipage}[b]{0.49\linewidth}
% 		\centering
% 		\includegraphics[width = .99\columnwidth]{overlap4.jpg}
% 		\centerline{\scriptsize(\textbf{b}) point cloud projection.} 
% 		\label{fig:pcp4}
% 	\end{minipage}
	\caption{Camera pose estimation visualization.(First row: NSH wall, Hamburg Hall windows; Second: Hamerschlag Hall, NSH lounge) }
	\label{fig:posevisualization}
\end{figure}

The final camera poses are shown in Fig.\ref{fig:posevisualization}. For each scene, we give an example of the true pose and our estimated pose on the point cloud map. The true poses are marked as $txyz$ in dash lines, while the estimations are $xyz$ in continuous lines. There is a small drift of camera positions, but the orientations of axes are parallel to each other. To better visualize the registration results, we project the original point cloud to the image plane with the same camera model and the estimated camera pose, which are shown on the top-right of each figure. The projected image is fused with the original camera image to visualize the misalignment. The overlapped area (brighter area) with small motion blur means that the registration result is better. From the global perspective, it overlaps well and the motion blur is minimal. When the depth changes dramatically, there exists some drift caused by misalignment.

{\color{black}To quantitatively analyze the results, we use the mean and standard deviation of 2D-3D matching pairs' number $Num$, rotation error $R_e (^\circ)$, position error $t_e (m)$, and position error related to the ground truth to measure the performance.} Tab.\ref{tab:dataset} shows the registration results for each scene with 10 images respectively. For NSH wall data, the total number of matching 2D-3D line segments is relatively smaller compared with other two outdoor scenes, but the line distributions are reliable. For Hamburg Hall windows, there exists more repeated structures, but the results are still feasible. There are more than 180 matching segments for each image.  For Hamerschlag Hall, there are both repeated structures and dramatic depth changes. The estimation errors are slightly bigger, 0.62 degrees for the mean rotation error and 0.54 meters for the mean position error. For NSH lounge, because it is an indoor environment, the total number of matching 2D-3D line segments is much smaller compared with outdoor scenes. Fortunately, the distance from camera to object is relatively small, we can get very high precision of pose estimation once sufficient 2D-3D line correspondences are found. {\color{black}We can also observe that there exist some disturbances for both rotation and position errors within acceptable ranges due to some negative factors (e.g. motion blur).}

\begin{table}[htbp]
	\caption{{\color{black}Matching quantities and registration errors for real data.}}
	\centering
	\setlength{\tabcolsep}{1.0mm}{
		\small
		\begin{tabular}{c|cc|cc|cc|cc}
			\hline
		    \multirow{2}{*}{$scene_{id}$} & \multicolumn{2}{|c}{\text{NSH wall}} & \multicolumn{2}{|c}{\text{Hamburg}}& \multicolumn{2}{|c}{\text{Hamerschlag}} & \multicolumn{2}{|c}{\text{NSH lounge}} \\
			  & mean & std & mean & std &mean & std &mean & std \\ \hline
			Num  & 91.50   & 15.72  & 183.80  & 13.46 & 183.50 & 23.62 & 16.100 & 2.73    \\ 
			$R_e(^\circ)$ & 0.36 & 0.18 & 0.65 & 0.32 & 0.62 & 0.27 & 0.20 & 0.13    \\
			$t_e(m)$   & 0.20 & 0.12 & 0.39 & 0.17 & 0.54 & 0.27 & 0.15 & 0.06   \\
			$t_e/t_{GT}$ &  0.04 & 0.02 & 0.06 & 0.02 & 0.17 & 0.09 & 0.10 & 0.04 \\
			\hline
	\end{tabular}}
	\label{tab:dataset}
\end{table}  

{\color{black}
For the testing on large scale environments, we extend our method to match a single image with a large scale point cloud map. When a pose estimation is obtained, we further check the visibility of 3D line features and discard the invisible 3D line matching based on \cite{biasutti2018visibility}. The performances on two large scale maps are shown in Fig.\ref{fig:largescale}. Both the position errors are less than 0.5 meters and the rotation errors are less than 1 degree.  To visualize the registration results, we further use the RGB image and camera pose to reproject the RGB information to the point clouds, so the visible point clouds are colorized. For Fig.\ref{fig:largescale}(b), we reproject the color information of two non-overlapping images to point clouds using the estimated camera poses. We can observe that there is no artifact in these areas, which means the localization is accurate.
\begin{figure}[h]
	\centering
	\begin{minipage}[b]{.49\linewidth}
		\centering
		\includegraphics[width =1.0\columnwidth]{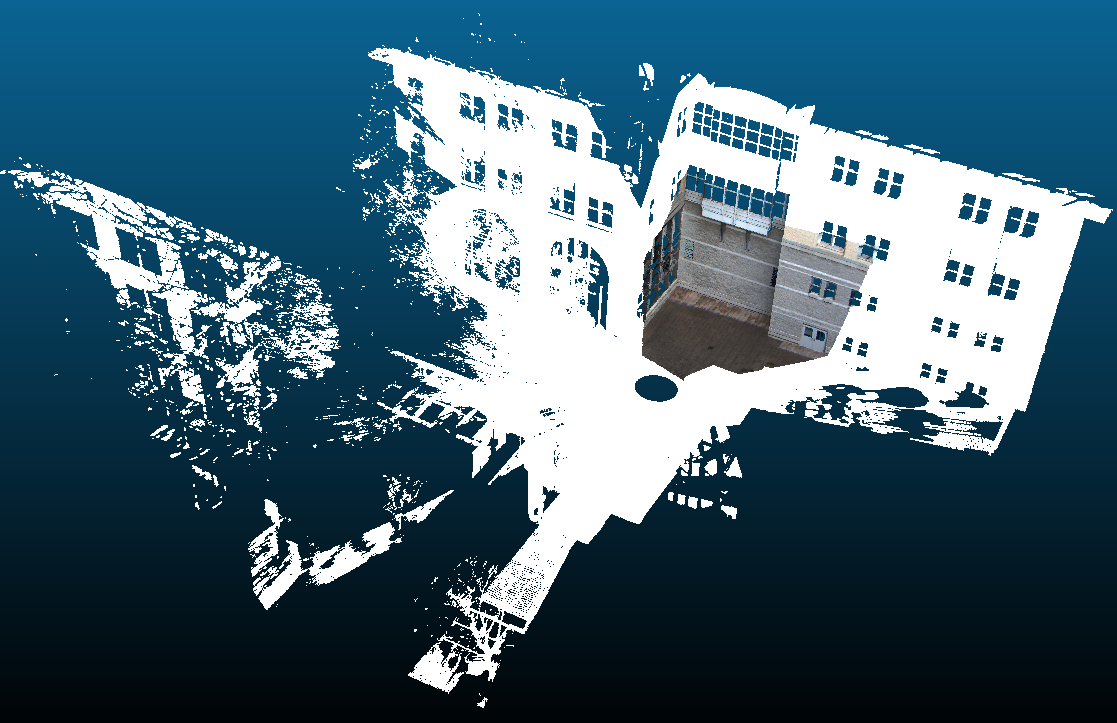} 
		\centerline{\scriptsize(\textbf{a}) NSH wall } 
		\label{fig:nsh}
	\end{minipage}
	\begin{minipage}[b]{.49\linewidth}
		\centering
		\includegraphics[height =1.1 in, width =1.0\columnwidth]{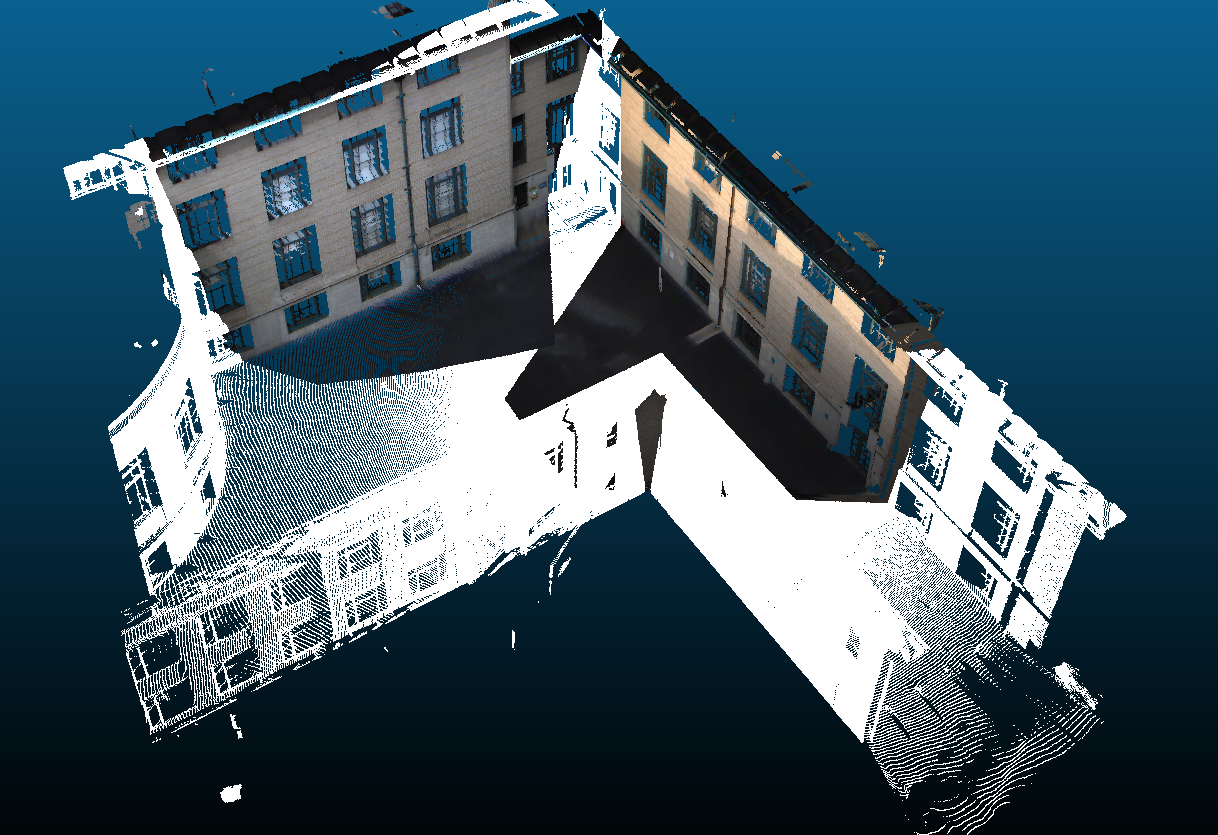}
		\centerline{\scriptsize(\textbf{b}) Hamburg Hall windows} 
		\label{fig:3dline}
	\end{minipage}
	\caption{ Colorization of large scale point clouds using estimated camera poses.}
	\label{fig:largescale}
\end{figure}
}

Regarding the processing time, the 3D line segment extraction costs the most time, which is related to the volume of point cloud. With the extracted 3D line segments, it takes about 8 seconds to estimate the camera pose of a single image on the Matlab implementation using an 8-core Intel i7 CPU. However, the time changes a little depending on the number of merged 2D and 3D line segments. We use parallel computing for different rotation candidates, thus it does not add much time for the estimation of translation vectors. 

During the experiments, we recommend to set the clusters of the vanishing point and 3D parallel line as 5. In Manhattan world, the common vanishing point number is 3. But 3 clusters sometimes result in 2 or 1 vanishing points after merging. Therefore, 5 clusters can yield more stable and robust output. Another parameter is the overlap length threshold. It is a valid 2D-3D correspondence when the overlap length exceeds half of the 2D line length. This is an empirical setting based on the performance. A larger setting can reject more inliers while smaller yields more outliers. This setting suits well both in the synthetic and real data experiments. In general, the estimated poses are promising, especially for the rotation calculation being less than 1 degree error for different scenes. The translation error may change a lot with structure repetitions and depth changes. Meanwhile, if we use RANSAC from the beginning with 6 line correspondences to estimate both $R$ and $t$, it rarely succeeds. This is because the search can easily fall into a local optimum. In our proposed method, the strategy using vanishing direction matching and hypothesis testing greatly reduces the chance of getting stuck into a local optimum. 

\section{Conclusion}
\label{sec5}
In this paper, we have presented an image to point cloud registration method to simultaneously estimate line correspondence and camera pose in structured environments. Based on geometric information, the method decouples the rotation calculation and translation estimation in two steps. Eight rotation candidates are obtained using the correspondence of vanishing directions to 3D primary parallel directions. Then a hypothesis testing approach is used to estimate the line correspondence and translation vector. Specifically, the framework using the hypothesis testing approach successfully deals with the rotation ambiguity from matching vanishing directions with 3D orientations. Experiments were conducted on synthetic and real data (both outdoors and indoors) with challenging repeated structures and rapid depth changes. The results demonstrate that the proposed method can effectively estimate line correspondence and camera pose.  In the future, we will exploit more efficient ways to find the global solutions of camera pose using line correspondence.

%\addtolength{\textheight}{-12cm}   % This command serves to balance the column lengths
                                  % on the last page of the document manually. It shortens
                                  % the textheight of the last page by a suitable amount.
                                  % This command does not take effect until the next page
                                  % so it should come on the page before the last. Make
                                  % sure that you do not shorten the textheight too much.

%%%%%%%%%%%%%%%%%%%%%%%%%%%%%%%%%%%%%%%%%%%%%%%%%%%%%%%%%%%%%%%%%%%%%%%%%%%%%%%%

\section*{ACKNOWLEDGMENT}
{\color{black}Sincere thanks are given to the anonymous reviewers and members of the editorial team for their comments and valuable recommendations. }
The authors also want to thank Warren Whittaker from CMU for the instructions on using FARO scanner and Dylan Campbell from ANU for the discussions and helps. %Huai Yu is supported by China Scholarship Council. 
%%%%%%%%%%%%%%%%%%%%%%%%%%%%%%%%%%%%%%%%%%%%%%%%%%%%%%%%%%%%%%%%%%%%%%%%%%%%%%%%
\bibliographystyle{IEEEtran}
\bibliography{egbib}
\end{document}